\documentclass[dvipsnames,format=sigconf,anonymous=false,review=false]{acmart} 
\usepackage{rotating,graphicx}
\usepackage{subcaption}
\usepackage{paralist}
\usepackage{url}
\usepackage{xcolor}
\usepackage{xspace}
\usepackage{graphicx}
\usepackage{url}
\usepackage[linesnumbered,ruled,noend]{algorithm2e}
\usepackage{bm}
\usepackage{booktabs}
\usepackage{array,multirow}
\usepackage{hyperref}

\newcommand{\lipiae}{Lipi-AE\xspace}
\newcommand{\lipiAE}{Lipi-AE\xspace}
\newcommand{\lexicase}{CONJUNCTIVE\xspace}
\newcommand{\lipiaesimple}{Lipi-AE-S\xspace}

\newcommand{\sizeMetric}{preserved percentage\xspace}
\newcommand{\randomMu}{RANDOM\xspace}
\newcommand{\MuVariance}{VARIANCE\xspace}
\newcommand{\MuLexicase }{CONJUNCTIVE\xspace}

\newcommand{\horse}{\texttt{Lipizzaner}\xspace}

\newcommand{\lipi}{\horse}

\newcommand{\alg}{Alg.\xspace}

\newcommand{\ns}{\mathbf{n}}

\SetKwProg{ParFor}{parfor}{}{}

\AtBeginDocument{%
  \providecommand\BibTeX{{%
    \normalfont B\kern-0.5em{\scshape i\kern-0.25em b}\kern-0.8em\TeX}}}

\copyrightyear{2025} 
\acmYear{2025} 
\setcopyright{cc}
\setcctype{by}
\acmConference[GECCO '25]{Genetic and Evolutionary Computation Conference}{July 14--18, 2025}{Malaga, Spain}
\acmBooktitle{Genetic and Evolutionary Computation Conference (GECCO '25), July 14--18, 2025, Malaga, Spain}\acmDOI{10.1145/3712256.3726449}
\acmISBN{979-8-4007-1465-8/2025/07}

\usepackage{algorithmicx}

\DeclareMathOperator*{\argmin}{arg\,min}

\usepackage{balance}

\begin{document}

\title[]{Guiding Evolutionary AutoEncoder Training with Activation-Based Pruning Operators}




\author{Steven Jorgensen}
\email{stevenjson@mit.edu}
\orcid{0000-0002-7782-0969}
\affiliation{%
\country{MIT, USA}
}

\author{Erik Hemberg}
\email{hembergerik@csail.mit.edu}
\orcid{0000-0002-2153-3506}
\affiliation{%
\country{MIT, USA}
}

\author{Jamal Toutouh}
\email{jamal@uma.es}
\orcid{0000-0001-6923-8445}
\affiliation{%
  \country{ITIS UMA, University of Malaga, Spain}
}

\author{Una-May O'Reilly}
\email{unamay@csail.mit.edu}
\orcid{0000-0003-1152-0346}
\affiliation{%
\country{MIT, USA}
}

\renewcommand{\shortauthors}{Jorgensen, et al.}


\begin{abstract}
This study explores a novel approach to neural network pruning using evolutionary computation, focusing on simultaneously pruning the encoder and decoder of an autoencoder. We introduce two new mutation operators that use layer activations to guide weight pruning. Our findings reveal that one of these activation-informed operators outperforms random pruning, resulting in more efficient autoencoders with comparable performance to canonically trained models. Prior work has established that autoencoder training is effective and scalable with a spatial coevolutionary algorithm that cooperatively coevolves a population of encoders with a population of decoders, rather than one autoencoder. We evaluate how the same activity-guided mutation operators transfer to this context. We find that random pruning is better than guided pruning, in the coevolutionary setting. This suggests activation-based guidance proves more effective in low-dimensional pruning environments, where constrained sample spaces can lead to deviations from true uniformity in randomization. Conversely, population-driven strategies enhance robustness by expanding the total pruning dimensionality, achieving statistically uniform randomness that better preserves system dynamics. We experiment with pruning according to different schedules and present best combinations of operator and schedule for the canonical and coevolving populations cases.

\end{abstract}




\maketitle

\section{Introduction}
\label{sec:introdiction}
A question that remains open in deep learning is why deep models need to be trained at a greater capacity, i.e. with many more parameters, than the dimensionality of their inputs.  
This question is important because the high capacities of deep networks make them very expensive to train and still quite expensive to use for inference.
Examining the puzzle of high capacity training leads to a related question, can the parameters that are not integral to effective learning and inference be identified and pruned away?  
 The Lottery Ticket Hypothesis, introduced within a seminal paper by Frankle et al. ~\cite{frankle2018lottery} provides insight into these questions. 
The paper finds that only a subset of parameters over the course of training become integral to the network's performance, but some redundant parameters seem to be necessary at one point or another, however, for training to be successful. Despite this observation, the underlying reasons for this phenomenon remain unclear.

Thus, while pruning can be applied,  achieving effective performance with it is not a simple  proposition.  
Digging into the detailed questions of when and how to prune has led to a variety of approaches to pruning, including EC-based contributions (see Section~\ref{sec:related_work}).  
In contrast to existing research, our approach simultaneously prunes a cooperating pair of neural networks, the encoder and decoder of an autoencoder. 

An autoencoder (AE) is an important model in the arsenal of deep learning models because it generates compact representations of unlabeled data, using unsupervised training \cite{Goodfellow-et-al-2016,ng2011sparse}.
While pruning has been extensively studied in convolutional neural networks (CNNs) and classification tasks, its application to AEs and clustering tasks remains relatively unexplored~\cite{altmann2024findingstronglotteryticket,Wu2021DifferentialEB,Wang2021EvolutionaryMM}. 

Our central research question is \textit{whether evolution-based pruning of an autoencoder (AE) can be effective}? 
To start, we introduce three different mutation operators that select AE parameters to prune. 
One operator is simple:  with a small probability, \textbf{Random Weight Pruning (\randomMu)} selects a parameter. It is applied to every parameter of the network.
Our other two activation-guided operators have heuristics for choosing whether to prune a parameter. They differ first on whether held-out or training data is used as input.  
Our first activation guided operator, \textbf{Variance Pruning (\MuVariance)} sums the activations across training inputs and then selects whether a parameter is pruned or not based on the variance of the sample. 
This approach is similar to that used to identify circuits in deep networks when using a mechanistic interpretability~\cite{conmy2023towards}.
Our second activation-guided operator, \textbf{Boolean Conjunctive Pruning (\MuLexicase)} avoids the large scale input data summation of \MuVariance and instead uses a heuristic that prunes based on similarity between input samples.

Our next research question introduces another dimension to explore for effective pruning - \textit{when should pruning occur and how should it be scheduled?} 
We test multiple schedules, which all prune the networks during training.
Our schedules include fixed interval pruning, linearly and exponentially changing pruning probability over time, and pruning only near the end of training. 
Addressing the two axes of what and when to prune with our operators and schedules allows us to experiment with mixing them, obtaining a set of pruning combinations. 
To evaluate the combinations, our metric for performance is  out-of-sample loss, and we also examine the pruned size of the final network relative to the original capacity, what we call the ``\sizeMetric".  We find many combinations that are more effective than not pruning and details follow in Section~\ref{sec:results--analysis} where we explicitly compare pruning combinations.

\begin{figure}

  \centering
    \captionsetup{font=footnotesize}
  \includegraphics[width=0.48\textwidth]{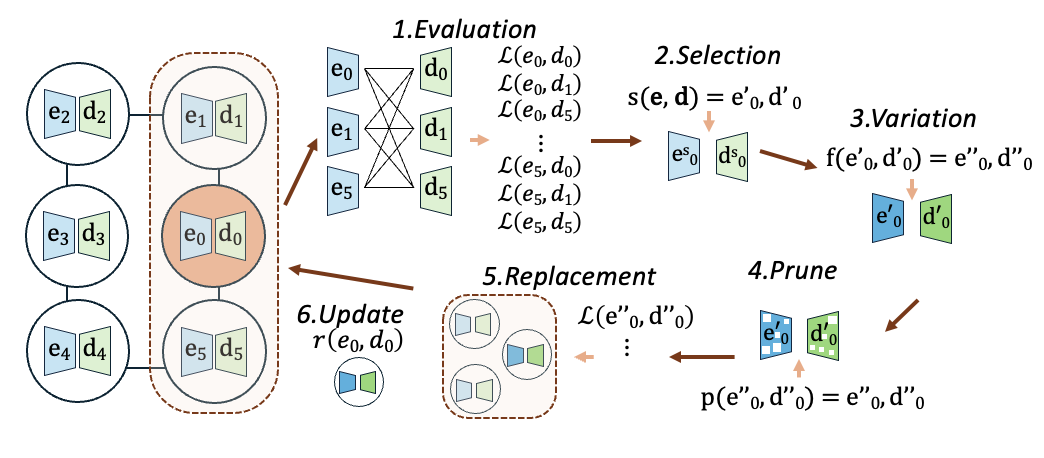}
  \caption{Overview \lipiae. A spatial ring topology of size 6 with
    radius one is shown. The neighbors are copied to form a
    subpopulation. The Euclidean product of AEs are evaluated. High quality encoders and decoders are selected. Then
    encoders and decoders are varied through training and
    replaced. Finally, the Euclidean product of AEs are evaluated and
    the encoder and decoder at the node are updated by the new best
    encoder and decoder.}
  \label{fig:overview_lipi_ae}
\end{figure}

We note that these experiments so far each prune only one autoencoder. 
Our final research question asks ``\textit{How well do the combinations of mutation operator and schedule transfer when using coevolution on a clustering problem?}". Further, how do they compare to one another in terms of performance and \sizeMetric? We turn to prior work, \lipiae, which applies evolutionary algorithms to AE training and shows promise in improving model resilience~\cite{Hemberg2024ae}.
\lipiae is a spatial coevolutionary  algorithm that cooperatively coevolves a population of encoders with a population of decoders in order to train a high performing encoder. 
Figure~\ref{fig:overview_lipi_ae} contains an overview of the algorithm.
With \lipiae, the finding of  Frankle et al. ~\cite{frankle2018lottery} along with our evolutionary operator and schedule combination finding raises intriguing questions about the interplay between evolutionary training and pruning in the context of AEs.
Our evaluation is conducted with a modestly simplified version of \lipiae which we call \lipiaesimple.

To summarize, our contributions are as follows:
\begin{asparaitem}
\item Two novel activation-guided mutation operators for pruning that differ with respect to the input data they reference and their approach to selecting which parameters to prune.  
\item A comparative analysis of the combinations of pruning operators, for single (a.k.a canonical) AE training and a determination of which is best for clustering. For canonical AE pruning \lexicase with an exponential pruning schedule performs competitively with the unpruned network.
\item A comparative analyses of the combinations of pruning operators for population-based, spatially distributed coevolutionary training. We observe that \randomMu with an exponential pruning schedule performs the best, even outperforming the canonical pruned networks.
\item An analysis of the interplay between pruning schedules and performance. We find that pruning schedules that prune with higher probability later in the training, such as exponential or final-$n$ schedule, have the highest performance.
\item A comparative analysis of encoder and decoder pruning patterns in \lipiaesimple and Canonical Autoencoders (AEs) using optimal pruning method-schedule combinations. We note pruning disparity between encoders and decoders in Canonical AEs, yielding smaller overall networks compared to \lipiaesimple AEs, which are more evenly pruned for two of the three methods.
\end{asparaitem}

\section{Background}\label{sec:prelims}

\subsection{ANNs, Lottery Ticket Hypothesis, and Pruning}
We adopt notation from~\cite{Hemberg2024ae}, where an ANN is defined as a parameterized function $f_{\theta}: \mathcal{X} \rightarrow \mathcal{Y}$ with $y = f_{\theta}(x)$. Parameters $\theta \in \mathbb{R}^n$ are optimized through iterative gradient descent updates:
$$\theta_t = \eta \mathcal{L}(y, f_{\theta}(x)) + \theta_{t-1},$$
where $\eta \in \mathbb{R}$ denotes the learning rate and $\theta_0$ is initialized from a uniform distribution $\mathcal{U}$. 
The loss function $\mathcal{L}: \mathcal{X} \times \mathcal{Y} \rightarrow \mathbb{R}$, $\mathcal{L}(f_{\theta}(x), y)$ quantifies prediction error, with updates computed over mini-batches of size $c$, $0 \leq c \leq |X|$ from dataset $D = [(x_0, y_0), \dots]$, where $|X|/c$ determines the update frequency. 
The objective of ANN training is to minimize some
loss, $$\argmin_{\theta \in \Theta} \mathcal{L}(y, f_{\theta}(x| D))$$. 


ANNs are often over-parameterized, i.e., they have more capacity in the form
of parameters to represent the function they are used to
approximate. This surplus representational capacity~\cite{Goodfellow-et-al-2016} allows ANNs to achieve high performance but comes at the cost of inefficiency in terms of computational and memory resources. Pruning addresses this issue by removing redundant parameters while preserving the network's accuracy~\cite{han2015learning}, which improves execution resource efficiency. 
The Lottery Ticket Hypothesis (LTH)~\cite{frankle2018lottery} suggests that within a randomly initialized, over-parameterized network there exist sparse subnetworks— i.e. ``winning tickets" —that are easier to train. These subnetworks, when trained in isolation, can achieve performance comparable to the fully trained original network. Furthermore, the hypothesis implies that if such subnetworks could be identified early, it might be possible to bypass the conventional training processes altogether.

Pruning implementations require careful consideration of multiple design aspects. The granularity of pruning determines structural impact: structured approaches remove entire filters, channels, or layers, while unstructured pruning targets individual weights for finer granularity~\cite{han2015learning}. Scheduling strategies dictate when pruning occurs; options range from one-shot (pre-training, post-training) to iterative protocols that alternate pruning with retraining phases~\cite{han2015learning,frankle2018lottery}. The pruning quantity, specifying the percentage of parameters to remove, serves as an important hyperparameter governing pruning intensity. Parameter selection methodologies further diversify implementations: magnitude-based approaches eliminate weights with smallest absolute values~\cite{frankle2018lottery}; activation-based methods prune neurons demonstrating minimal activation patterns during inference~\cite{liurethinking,sunsimple}; and low-rank approximation techniques decompose weight matrices into compact factorized representations~\cite{han2015learning}.

In this contribution, we indirectly address pruning quantity through the pruning operators which are activation based (we refer to them as activation-guided) and explore different schedules.


\subsection{Autoencoders}
\label{sec:autoencoder}

An autoencoder (AE) consists of two learned functions: an encoder $e_{\theta_e}: \mathcal{X} \rightarrow \mathbb{R}^n$ that maps inputs $\mathbf{x} \in \mathcal{X}$ to latent representations $\mathbf{z} \in \mathbb{R}^n$, and a decoder $d_{\theta_d}: \mathbb{R}^n \rightarrow \mathcal{X}$ that reconstructs inputs from latent codes~\cite{b2012}. 
The full transformation is $\mathbf{x}' = d_{\theta_d}(e_{\theta_e}(\mathbf{x}))$ where $\theta_e$ and $\theta_d$ denote encoder and decoder parameters. The primary training objective minimizes the reconstruction error:
\begin{equation*}
\argmin_{\theta_e, \theta_d} \mathbb{E}{\mathbf{x} \sim D} \left[ \mathcal{L}_{ae}(\mathbf{x}, d_{\theta_d}(e_{\theta_e}(\mathbf{x}))) \right]
\end{equation*}
with the base loss typically being $\mathcal{L}_{ae}(\mathbf{x}, \mathbf{x}') = |\mathbf{x} - \mathbf{x}'|$. Parameter updates follow simultaneous gradient descent:
\begin{align}
\theta_{t,e} &= \theta_{t-1,e} - \eta \nabla_{\theta_e}\mathcal{L}_{ae} \
\theta{t,d} &= \theta_{t-1,d} - \eta \nabla_{\theta_d}\mathcal{L}_{ae}
\end{align}

During training, practitioners often monitor loss dynamics to detect optimization issues:
\begin{equation*}
\exists C \in \mathbb{N} \text{ s.t. } \mathcal{L}_{ae}^{(t-C)} < \mathcal{L}_{ae}^{(t)} \text{ for consecutive iterations } t-C,...,t
\end{equation*}

This degradation signal typically triggers early stopping or learning rate adaptation.

\subsection{Mechanistic Interpretability}
\label{sec:mech-interpr-anns}

\begin{figure}[]
  \centering
  \begin{subfigure}{0.23\textwidth}
  \captionsetup{font=footnotesize}
  \includegraphics[width=0.99\textwidth]{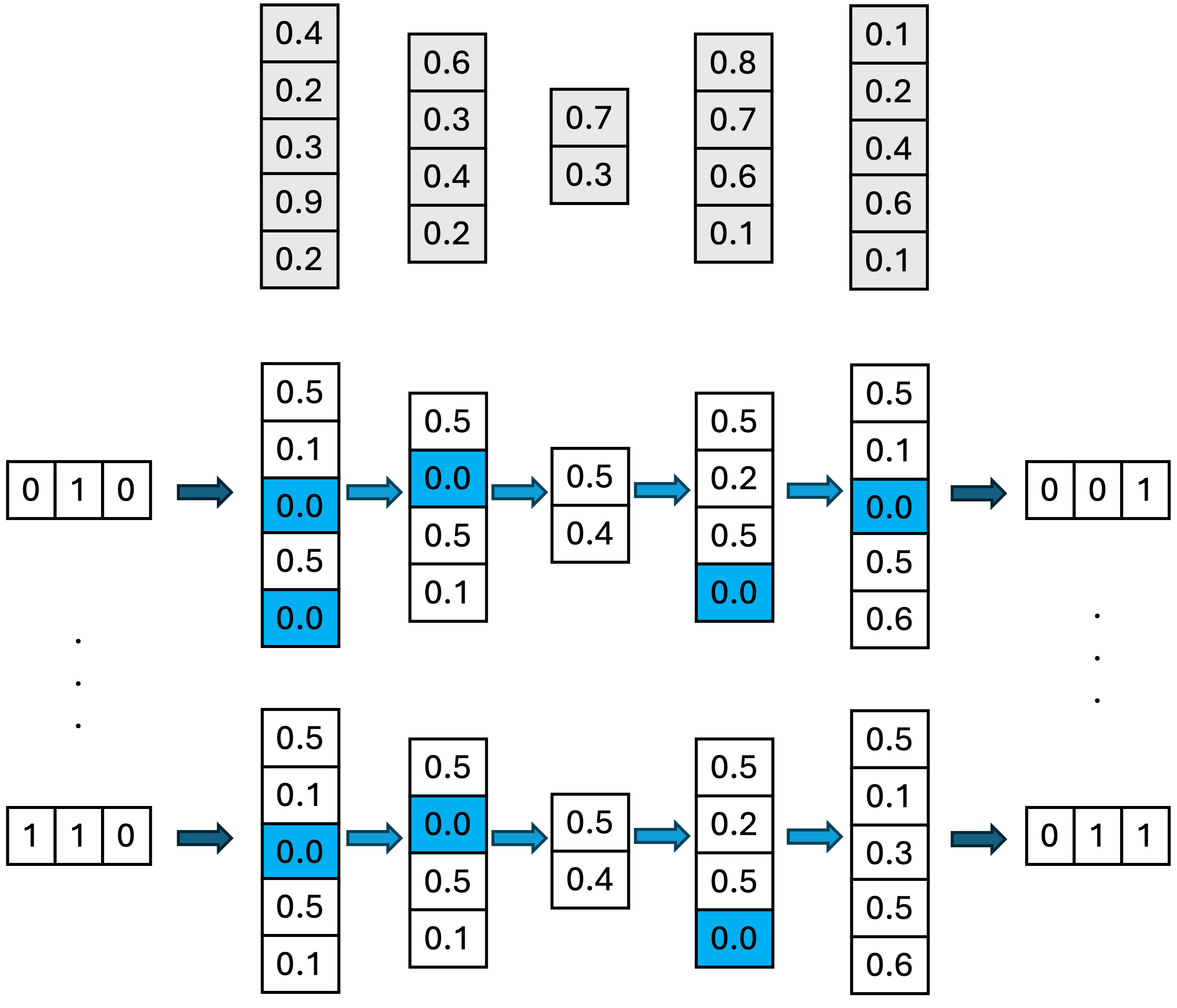}
  \caption{Original ANN}
  \label{fig:mi_overview_org}
\end{subfigure}
\begin{subfigure}{0.23\textwidth}
  \centering
    \captionsetup{font=footnotesize}
  \includegraphics[width=0.99\textwidth]{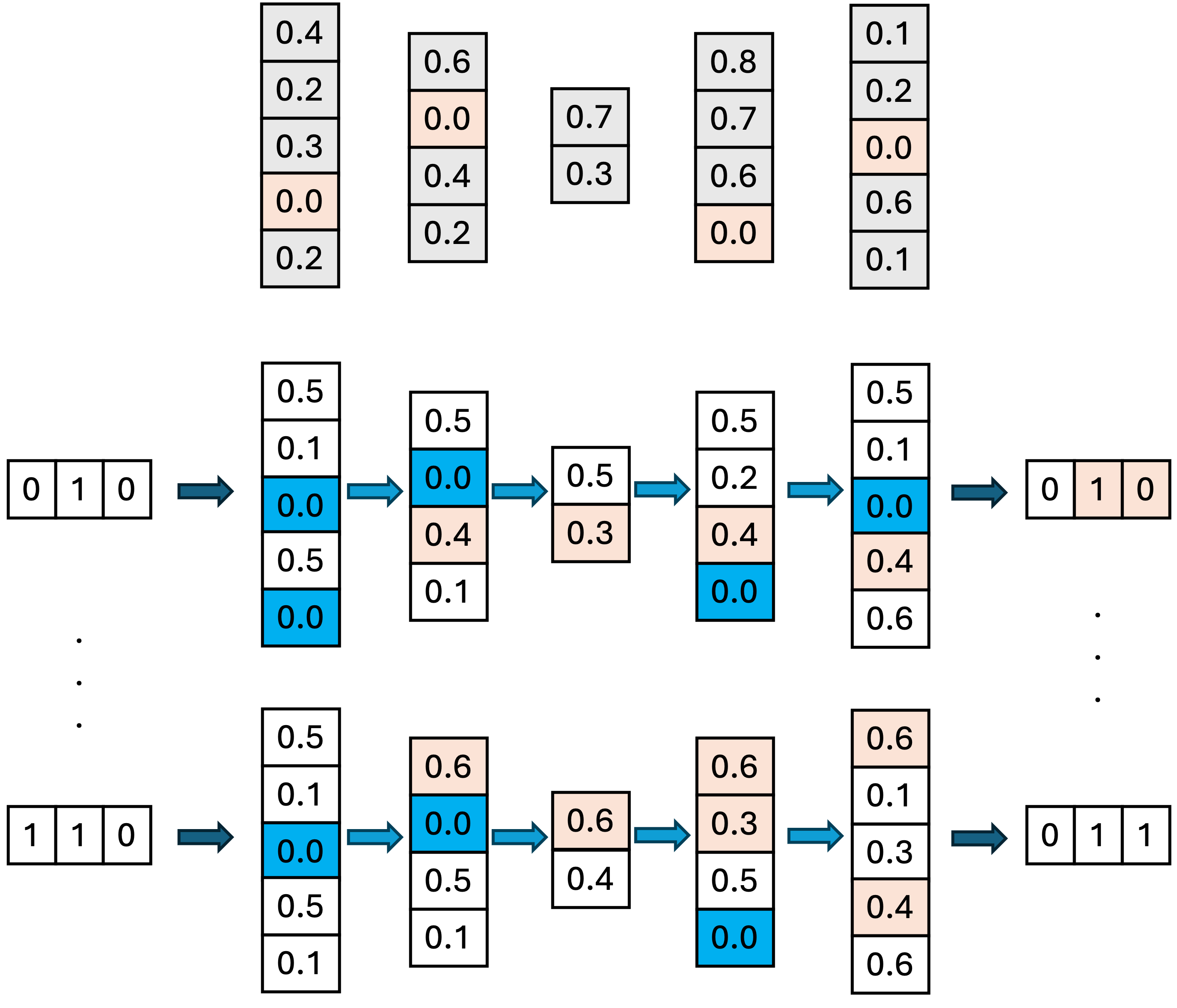}
  \caption{Pruned ANN}
  \label{fig:mi_overview_pruned}
\end{subfigure}
  \caption{Overview of mechanistic interpretation of an AE
    compute graph that is fully connected. Gray illustrates node weight parameters. Blue
    illustrates low activations for input data which leads to
    different paths, based on activation at
    node. Red illustrates changes in parameter and activation between
    original ANN~\ref{fig:mi_overview_org} and pruned
    ANN~\ref{fig:mi_overview_pruned}}
  \label{fig:mi_overview}
  \vspace{-0.5cm}
\end{figure}

Behavioral analysis of ANNs, often referred to as mechanistic interpretability (MI) for vision models and transformer-based models~\cite{elhage2021mathematical}, can inform design improvements and foster trust~\cite{meng2022locating, conmy2023towards}. This involves reverse-engineering ANN behavior by analyzing activations on carefully selected inputs using invasive and non-invasive techniques~\cite{conmy2023towards, merullo2023mechanism}. Given input data $\mathbf{x}$ and function $f:\mathbb{R}^{n \times m}\rightarrow \mathbb{R}$, the activation values $A = f_{\theta}(\mathbf{x})$ describe the ANN's behavior, where $A_{i,j} \neq 0$ indicates active nodes in the compute graph. Figure~\ref{fig:mi_overview} illustrates MI for an AE compute graph, highlighting low activation paths (blue) and node weights (gray). The hypothesis behind activation-guided pruning is that low-activation paths or weights are more suitable for removal than high-activation counterparts.



\subsection{Coevolutionary Algorithms}\label{sec:coevalgs}

Biological coevolution refers to the mutual influence that two or more
species have on each other's evolutionary
processes~\cite{rosin1997new}. Coevolution can be
categorized as either cooperative, where there is mutual benefit, or
competitive, involving contested resources. In evolutionary algorithms
(EAs), individual solutions like fixed-length bit strings are
typically evolved, as in Genetic Algorithms (GAs)~\cite{Goldberg1989},
where a pre-determined fitness function evaluates the quality of
individuals. Conversely, coevolutionary algorithms (CAs) assess an
individual's fitness based on its interactions with members of a
second population, thus imitating species-to-species
interactions, see for example~\cite{Antonio2018,popovici2012,sims1994evolving,mitchell2006coevolutionary,krawiec2016solving,hemberg2016detecting,o2020adversarial}.


Applied to deep learning, CAs train adversarial (e.g., GANs) and cooperative (e.g., autoencoders) architectures through dual-population dynamics, enhancing model robustness and mitigating common training pathologies. Spatial coevolution frameworks like \horse~\cite{Hemberg2021spatial,Hemberg2024ae} address computational complexity through grid-based population: paired generator/encoder and discriminator/decoder subpopulations occupy overlapping cellular neighborhoods, enabling localized interactions that approximate full population engagement at reduced cost. Key \horse components are (1) gradient-based ANN parameter updates, (2) Gaussian mutations for hyperparameter adaptation, and (3) selection mechanisms preserving convergence while maintaining diversity.
\section{Related Work}
\label{sec:related_work}

\begin{table*}[]
  \centering
  \footnotesize
  \caption{Contributions Related to Evolutionary Deep Model Training.  Where the algorithm type is not clear from the publication, the table entry states EA.}
  \label{tab:rw_overview}
  \begin{tabular}{l|p{8.95cm}|l|l|p{1.5cm}|l}
    \textbf{Auth.} & \textbf{Title} & \textbf{Alg.} & \textbf{Activation-based?} & \textbf{Model} & \textbf{Task}\\
    \hline
    \cite{altmann2024findingstronglotteryticket} & Finding Strong Lottery Ticket Networks with Genetic Algorithms & GA & Yes & single ANN & Classification\\
    \hline
    \cite{Wu2021DifferentialEB} & Differential Evolution Based Layer-Wise Weight Pruning for Compressing Deep Neural Networks & DE & Yes & Single CNN & Classification\\
    \hline
    \cite{Wang2021EvolutionaryMM} & Evolutionary Multi-Objective Model Compression for Deep Neural Networks & MOO & Yes & Single ANN & Classification\\
    \hline
    \cite{VelayuthamC2023EvoPrunerPoolAE} & EvoPrunerPool: An Evolutionary Pruner using Pruner Pool for Compressing Convolutional Neural Networks & EA & Yes & Single ANN  & Classification\\
    \hline
    \cite{Chung2024MultiobjectiveEA} & Multi-objective evolutionary architectural pruning of deep convolutional neural networks with weights inheritance & MMO & Yes & Single ANN & Classification\\
    \hline
    \cite{Poyatos2022EvoPruneDeepTLAE} & EvoPruneDeepTL: An Evolutionary Pruning Model for Transfer Learning based Deep Neural Networks & EA & Yes & Single ANN & Classification\\
    \hline
    \cite{Wang2020NetworkPU} & Network pruning using sparse learning and genetic algorithm & GA & Yes & Single ANN & Classification\\
        \hline
        \cite{Hemberg2024ae} & Cooperative Spatial
Topologies for Autoencoder Training & CA & None & None & Clustering\\
        \hline
        Ours & Guiding evolutionary autoencoder training with mechanistic interpretations of behavior & CA & Yes and No & bi-model AE & Clustering\\
  \end{tabular}
\end{table*}

\paragraph{Coevolutionary Deep Model Training}This contribution and \cite{Hemberg2024ae}  are closely related to \lipi, with the major
difference being they train  AEs with a loss minimization objective instead of
training a GAN with a minimax objective. Additionally, one AE is returned here and in \cite{Hemberg2024ae} while \lipi returns a mixture of generators.

\paragraph{EC Approaches to Deep Model Pruning}  Most EC approaches to pruning ANNs consider weights without considering how much they are activated when input data is passed through the network. 
The lottery ticket hypothesis motivated the pruning of weight of binary classifiers done with Genetic
Algorithms~\cite{altmann2024findingstronglotteryticket}. 
In EvoPruneDeepTL the last fully-connected layers are pruned to become sparse layers
with a genetic algorithm in order to improve transfer
Learning~\cite{Poyatos2022EvoPruneDeepTLAE}.
A GA is also used to prune the weights of a convolutional NN
in~\cite{Wang2020NetworkPU}. 
In contrast, Differential
Evolution is used to find the optimal pruning sensitivity set for each
layer with binary masks that are applied to structures of the network
with single weights or convolution filters\cite{Wu2021DifferentialEB}.

Network pruning can be recognized in a model compression approach that evolves a population of architectures with multi-objective
optimization\cite{Wang2021EvolutionaryMM}.  
Multi-objective evolutionary architectural pruning of deep convolutional neural
networks was investigated with structured filtered pruning of DCNN
weights in ~\cite{Chung2024MultiobjectiveEA}. 
In contrast, EvoPrunerPool is
an evolutionary approach to pruning that formulates CNN filter pruning
as a search problem to identify a set of pruners from a pool of
off-the-shelf filter pruners and applies them in sequence to
incrementally sparsify an ANN~\cite{VelayuthamC2023EvoPrunerPoolAE}.

Table~\ref{tab:rw_overview} summarizes and contrasts the different EC approaches to ANN pruning just mentioned.  Distinctions among them arise on different bases. One basis is the type of EA which includes genetic algorithms, multi-objective optimization, and differential evolution. A second is the type of deep model that is pruned. A third is the task the model is trained for. This contribution is novel on all three of these bases in its use of a coevolutionary algorithm to train a dual-model AE for a clustering task.  A final key basis is  whether pruning is importance-guided. Using activation-informed pruning is the final unique facet of this contribution.  For completeness, the table includes \cite{Hemberg2024ae} which trains AEs with cooperative coevolution on clustering, but does not use pruning.

\section{Method}
\label{sec:method}


\subsection{\lipiaesimple}
\label{sec:lipiaesimple}

Our method, \lipiaesimple, builds upon the cooperative coevolutionary framework \lipiae by introducing simplified parallel execution and integrated pruning. The \lipiaesimple framework employs a spatial coevolutionary approach, evolving populations of encoders and decoders within a ring topology comprising $Z$ cells. Each cell $n_k$, where $k$ ranges from $1$ to $Z$, maintains a local subpopulation of encoders $e_k$ and decoders $d_k$. These subpopulations are formed by gathering individuals from neighboring cells within a specified radius $r$, resulting in a neighborhood size of $s = 2r + 1$.

As illustrated in Fig.~\ref{fig:overview_lipi_ae}, the ring topology is arranged as a one-dimensional grid of $Z$ cells. Each cell contains a subpopulation of $(\mathbf{e}, \mathbf{d})$ pairs, including its own pair and those from neighboring cells within the radius $r$. Following each evolutionary generation, the best central $(\mathbf{e}, \mathbf{d})$ pair is updated and shared with neighboring cells. For example, in Fig.~\ref{fig:overview_lipi_ae}, six individuals form a ring topology with a neighborhood radius $r=1$. The shaded areas highlight the neighborhood of cell (0), demonstrating how updates propagate to cells (5) and (1). 

At each generation $t \in \{0, \dots, T\}$, each cell synchronizes with its neighbors by copying their current central autoencoder to form local subpopulations. The cooperative coevolution process consists of four main steps, as outlined in \alg~\ref{alg:gan-coev-simple}. First, each center autoencoder pair $(\mathbf{e}, \mathbf{d})$ undergoes gradient-based training using stochastic gradient descent (SGD) to optimize its parameters. The models are then evaluated based on their reconstruction loss, which measures their ability to reconstruct input data.

Next, $(\mathbf{e}, \mathbf{d})$ is pruned to reduce its complexity of the networks. Our pruning methods works by applying a binary mask over the model's parameters. This mask sets pruned weights to zero, but it does not physically remove them from the model. Typically in post-training pruning, these zeroed weights remain unchanged, as training has already concluded. However, since our approach does pruning during training, the zeroed weights can be updated and become non-zero again in subsequent training epochs due to gradient updates.

Finally, the selection function uses tournament selection to replace underperforming autoencoders with the best from their neighborhood, driving improvement across generations through synchronized updates. After all epochs have been completed, \lipiaesimple selects the globally optimal autoencoder $(\mathbf{e}, \mathbf{d})$ based on reconstruction loss. 
\setlength{\textfloatsep}{2pt}%
\begin{algorithm}
  \footnotesize
  \SetKwComment{Comment}{//}{}
  \DontPrintSemicolon
  \SetKwInOut{Input}{Input}  \SetKwInOut{Output}{Return}
  \caption{\lipiaesimple: Evaluate and select a new sub-population. Each mini-batch
    trains the selected encoder and decoder $e, d$.  
    Return the new cell neighborhood $\ns$. 
    $\theta_{COEV}$ includes $\tau$: Tournament size, $\mathbf{B}$: Input training dataset batches, $\beta$: Mutation probability, $\mathbf{n}$: Cell neighborhood sub-population, and $C_{p}$: Pruning mutation parameters.}
  \label{alg:gan-coev-simple}
  \Input{
    ~$\mathbf{n}~$: Cell neighborhood sub-population,~$\theta_{COEV}$~: Parameters for \lipiaesimple}
  \Output{
    $\mathbf{n}$: Updated cell neighborhood sub-population
  }
  $\mathbf{n}_{\delta} \gets$ mutate\_learning\_rate($\mathbf{n}_{\delta}, \beta$) \label{alg:mutate-learning-rate-line}\\
  $\phi \gets$ evaluate\_AEs($B, \mathbf{n}$) \\
  $\mathbf{e}, \mathbf{d} \gets$ tournament\_select($\mathbf{n}, \phi, \tau$) \label{alg:sel-line}\\
  \For {$B \in \mathbf{B}$}{ 
      $\nabla_{e}, \nabla_{d} \gets$ compute\_gradient($e, d, B$) \label{alg:compute-gradient-line-1-s}\\
      $e, d \gets$ update\_ANN($e, d, \nabla_e, \nabla_d, B$)  \label{alg:update-nn-line-1-s}\\
    }
  
  $\phi \gets$ evaluate\_AEs($B, \mathbf{n}$) \\
  $e, d \gets$ prune\_ANN($e, d, \nabla_e, \nabla_d, C_{p}$) \label{alg:prune-nn-line-1}\\
    $\mathbf{n} \gets$ replace\_center\_individuals($\mathbf{n}, \phi$) \label{alg:replace-center}\\
  \Return $\mathbf{n}$ 

\end{algorithm}



\subsection{Pruning based variation}
\label{sec:behav-based-vari}

Our coevolutionary algorithm (Algorithm~\ref{alg:gan-coev-simple}, line~\ref{alg:prune-nn-line-1}) incorporates strategic pruning of autoencoder (AE) parameters through two mechanisms: temporal scheduling (``when to prune") and selection criteria (``what to prune"). This approach allows for systematic exploration of architectural optimization dynamics.
\subsubsection*{Pruning Scheduling}
\label{sec:prune-sched}
To determine when to prune the AE we investigate 6 different pruning
schedules:
\begin{itemize}
\item [\textbf{Fixed:}] A fixed pruning probability, $p_p$ at each epoch,
  $p_p(t) = C, C \in [0, 1] , \forall t \in T$
\item [\textbf{Increase:}] The pruning probability, $p_p$ increases linearly
  each epoch, $p_p(t) = p_p(t-1) + C/T, t \in T, p_p(t=0) = 0$
\item [\textbf{Decrease:}] The pruning probability, $p_p$ decreases linearly
  each epoch, $p_p(t) = p_p(t-1) - C/T$
\item [\textbf{Population:}] The pruning probability, $p_p$ is divided by the
  sub-population~(neighborhood) size,   $p_p(t) = C/|\mathbf{n}|$
\item [\textbf{Exponential:}] The pruning probability increases exponentially
  each epoch, $p_p(t) = C (1 - e^{-2t/T})$
\item [\textbf{Final-$n$:}] The pruning probability, $p_p$, is fixed in the last $t_p$
  epochs, otherwise it is zero $p_p(t) = C \textnormal{ if } t > T - t_p \textnormal{ else } 0$ 
\end{itemize}

\subsubsection*{Pruning Selection}
In addition to determining when to prune, it is equally important to decide which parameters should be pruned. We investigate three distinct methods for parameter selection.

The first method, random parameter pruning (\textbf{\randomMu}), selects a proportion of parameters ($p_a \in [0,1]$) uniformly for removal. The second method, variance-weighted pruning (\textbf{\MuVariance}), leverages activation statistics from the training dataset. Activations are collected for each parameter across all training samples, and their variances are computed. Parameters with lower variance are deemed less critical and are assigned higher probabilities of being pruned. The normalized inverse variance is used to weight parameter selection probabilities, ensuring that $p_a$ proportion of parameters are removed based on their relative importance.

The third method, (\textbf{\MuLexicase}), generates activations from $h$ held-out samples. For each sample, activations are normalized before applying a threshold function. The thresholded activations are then combined using a logical AND operation across all held-out samples. Parameters with an output of 1 from this conjunction are pruned immediately, as they are considered redundant for AE performance. If no parameters meet this criterion, one held-out sample is removed iteratively until prunable parameters are identified or all samples are exhausted. In cases where no parameters can be pruned after exhausting all samples, no pruning occurs for that epoch. This iterative process ensures that only parameters deemed unnecessary across multiple contexts are removed. Algorithm~\ref{alg:conj-comparison} provides a more detailed description of this method.

\begin{algorithm}
  \footnotesize
  \SetKwComment{Comment}{//}{}
  \DontPrintSemicolon
  \SetKwInOut{Input}{Input}  \SetKwInOut{Output}{Return}
  \caption{\textbf{\MuLexicase}: Select parameters in a
    layer to prune based on thresholded activations from held-out samples.}
  \label{alg:conj-comparison}
  \Input{
    $A$: Layer activations from $h$ held-out samples, $C$: Threshold value}
  \Output{
    $\mathbf{s}$: Selected nodes to be pruned
  }
  $A' \gets$ normalize($A$) $< C$\\

  $s \gets \bigcap_{i=1}^{|A'|}A'[i, :]$
  
  $n_s \gets \sum s$\\
  
    \While{$n_s < 1$ \text{\textbf{\&}} $|A'| > 0$}{
        $r \gets $ generateRandomInteger($1$, $|A'|$)
      
        remove $A'[r, :]$ from $A'$
        
        $s \gets \bigcap_{i=1}^{|A'|}A'[i, :]$
  
        $n_s \gets \sum \mathbf{s}$\\
    }
  
  \Return $\mathbf{s}$ 

\end{algorithm}

\section{Experiments}
\label{sec:experiments}

In this section, we present an experimental study to investigate the effectiveness of EC in training and pruning AEs. Our experiments are implemented to address the main research questions stated in Section~\ref{sec:introdiction}. 


\subsection{Binary Clustering Problem}
\label{sec:binary_clustering_problem}

\begin{figure}[]
\setlength{\abovecaptionskip}{0pt}
\setlength{\belowcaptionskip}{0pt}
  \centering
    \captionsetup{font=footnotesize}
  \includegraphics[width=0.2\textwidth]{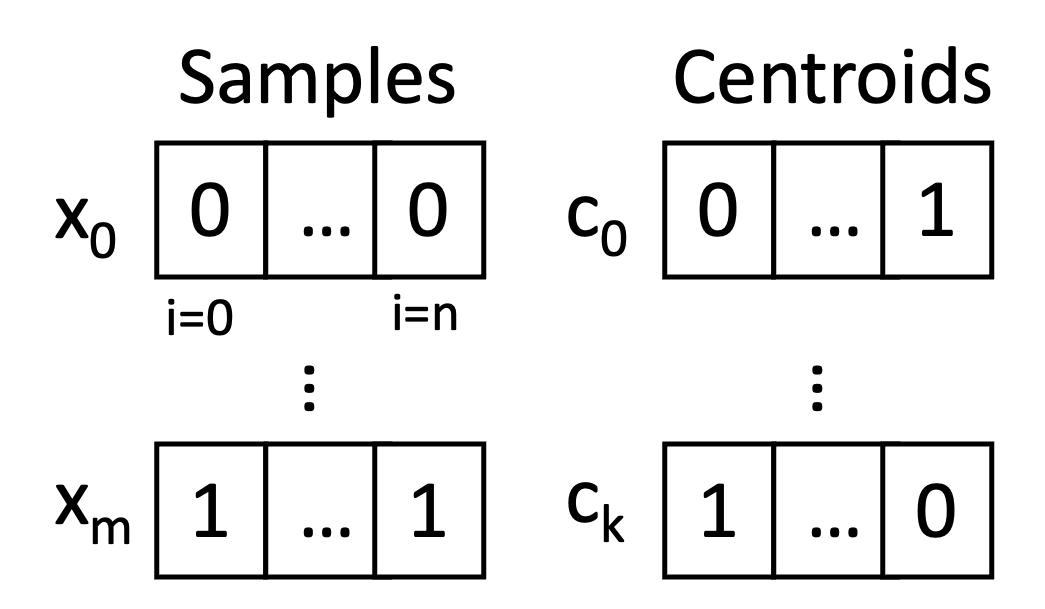}
  \caption{Illustration of the Binary Clustering Problem~\cite{Hemberg2024ae}. Sample vectors and centroids are shown.}
  \label{fig:bc_ex}
\end{figure}

We evaluate \lipiaesimple's reconstruction capabilities using a synthetic binary clustering task with tunable difficulty. The problem operates on $m$ input vectors $X = {x_1, \dots, x_m}$ where each $x_i \in {0,1}^n$, seeking to produce $k$ centroid vectors $C = {c_1, \dots, c_k}$ through a mapping function $f : {0,1}^{n \times m} \rightarrow {0,1}^{n \times k}$ that minimizes the cumulative Hamming distance:
\begin{equation*}
\min \sum_{t=1}^{m} \Delta(x_t, f(x_t))
\end{equation*}
where $\Delta$ denotes the Hamming distance between data points and their assigned centroids. This NP-complete problem~\cite{b2012} provides a controlled environment for analyzing autoencoder performance. An example of the problem is shown in Figure \ref{fig:bc_ex}.

We implement $n=1000$-dimensional binary vectors with $m=100$ samples and $k=10$ centroids. Centroids are initialized by independently sampling each bit from a Bernoulli distribution ($p=0.5$). Training examples are derived from these centroids through random bit flips applied with probability $q=0.05$, simulating real-world data corruption. Both training and test sets contain $nm$ samples.


\subsection{Setup}
\label{sec:setup}


Our experimental framework extends the CA training methodology established in \cite{Hemberg2024ae}, which identified optimal performance using the \lipiAE framework with best-solution selection. We implement 30 independent trials per configuration using L1 loss minimization, with default model architectural details available in Appendix~\ref{sec:results_appendix}.



\subsubsection*{Performance Metrics}
Network complexity is quantified via active (non-zero) parameter counts, while pruning effectiveness is measured by both the proportion of zero-weight parameters and task-specific accuracy. We employ test dataset loss as our primary performance metric, as preliminary analysis revealed nearly identical behavior between training and test losses (mean Pearson's $r = 0.97$). Activation patterns are tracked to characterize network behavior evolution.

\subsubsection*{Experimental Design}
Three fundamental research questions structure our investigation:

\begin{inparadesc}
\item \textit{Can evolution-based pruning of an AE be effective?} 
We contrast Canonical AE training and \lipiaesimple training. We use a default AE size of ~(61,670 parameters) for both methods and also try an AE with a smaller latent layer ~(61,182) (\lipiaesimple only). See Figure \ref{fig:base_ann_arch} in the Appendix for details on the default model architecture.
\item \textit{What is the impact of pruning schedule and pruning method on AE performance?}
Six pruning schedules are evaluated - fixed probability, increasing probability, population based probability, decreasing, exponential, and final-$n$. Schedule implementations follow the formal definitions in Section~\ref{sec:prune-sched}.
We investigate the following pruning methods (previously described in Section~\ref{sec:method}): None, VARIANCE, \MuLexicase, and RANDOM.
\item \textit{How do the methods compare in terms of preserved-percentage of the original network?}
We analyze and compare the size of the pruned network for each method after completing all epochs. Additionally, we assess the pruning progression over epochs, focusing on the pruning amount differences between the encoder and decoder components for each method.

\end{inparadesc}



\subsubsection*{Implementation Details}
All experiments use fixed hyperparameters ($\alpha = 10^{-5}$, batch size = 5) across $T=400$ training epochs (generations), selected through comprehensive preliminary sweeps (omitted for space). Our pruning methods leverage PyTorch's \textit{prune.remove} functionality, which zero's out pruned model weights rather than fully remove them from the architecture. The full code implementation can be found in our github repository\footnote{\url{https://github.com/ALFA-group/lipizzaner-ae-prune}}. Sensitivity analyses for population-based training appear in Appendix~\ref{sec:ablation-study}.



\subsection{Results \& Analysis}
\label{sec:results--analysis}
In this section, we present an analysis of our experimental results, shedding light on the effectiveness of various pruning methods and schedules in autoencoder training. 
Our findings reveal patterns in the interplay between pruning methods, schedules, and training approaches, which we discuss throughout this section. 

\subsubsection{Can evolution-based pruning of an AE be effective?}

\begin{figure}
    \centering
    \captionsetup{font=footnotesize}
    \includegraphics[width=0.95\linewidth]{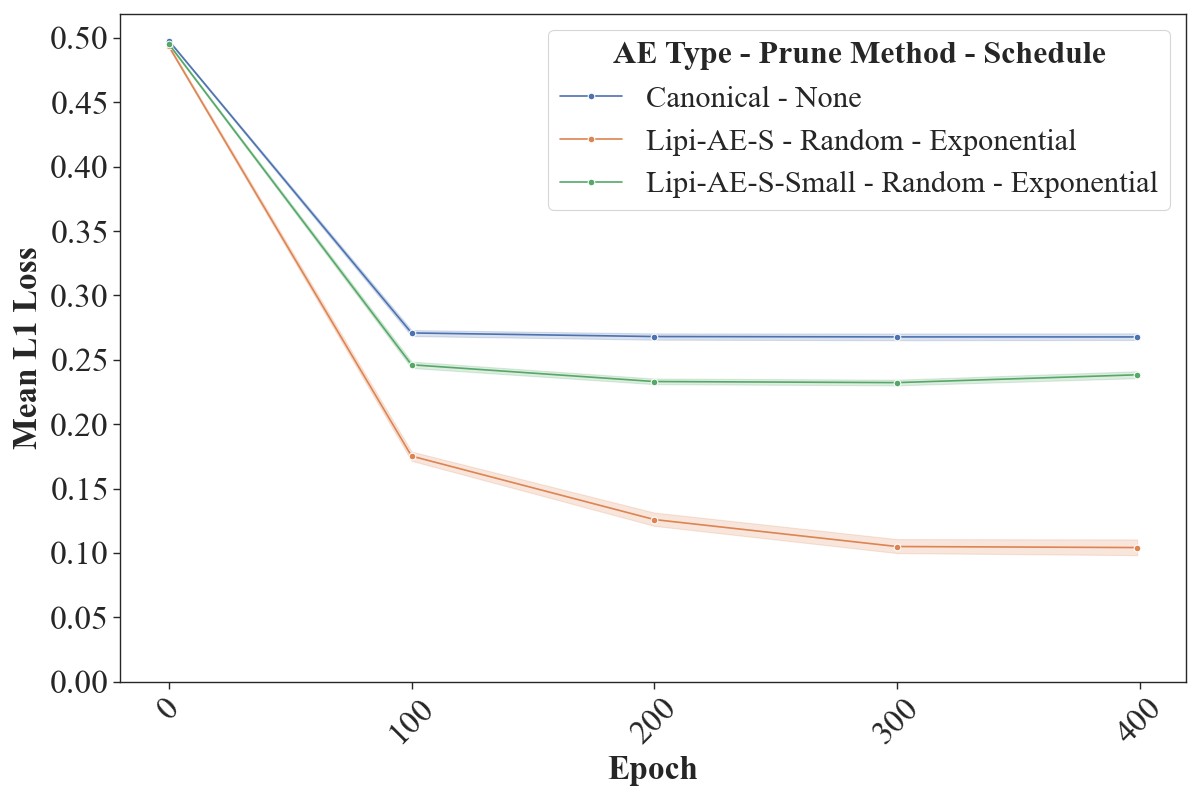}
    \caption{Best-performing combinations of pruning methods and schedules for \lipiaesimple training (default and reduced network sizes) compared to canonical AE training}
    \label{fig:best-all}
\end{figure}

\begin{figure*}[h]
    \captionsetup{font=footnotesize}
\begin{subfigure}{0.32\textwidth}
    \centering
    \captionsetup{font=footnotesize}
    \includegraphics[width=0.95\linewidth]{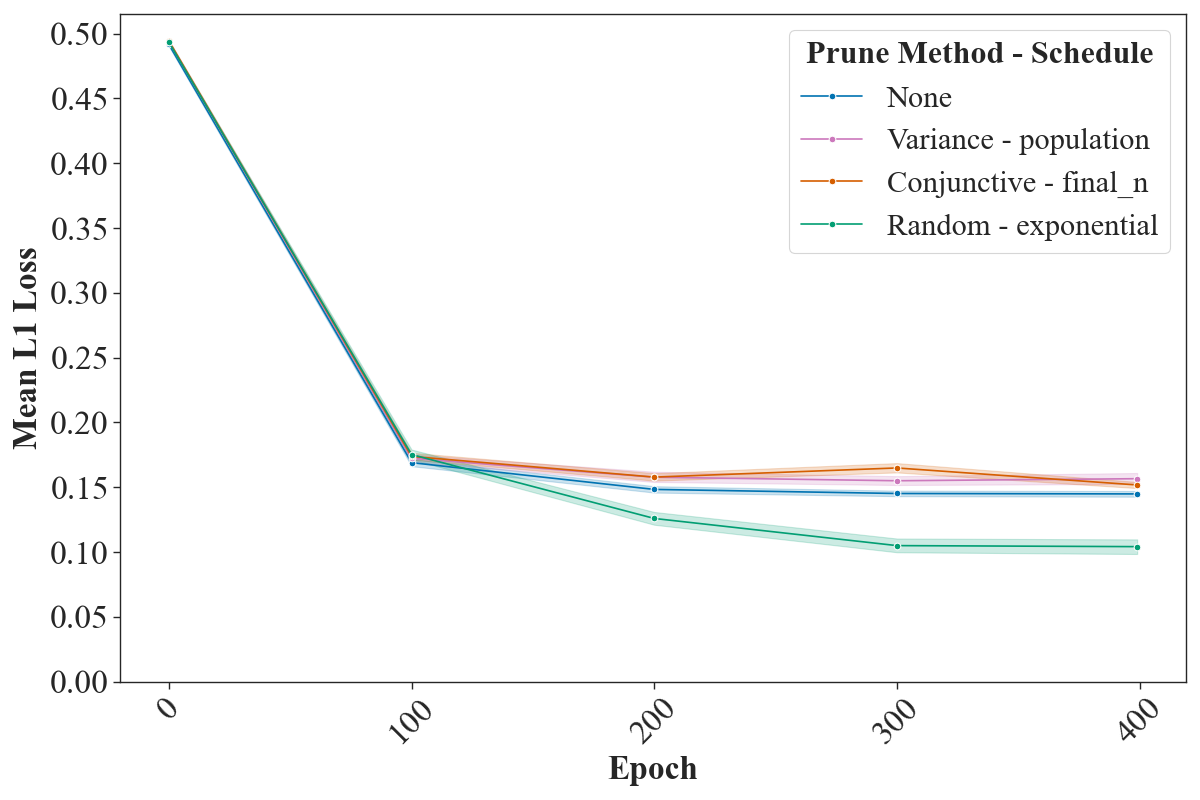}
    \caption{\lipiaesimple: Average L1 loss achieved by top-performing pruning schedules per method}
    \label{fig:best-lipi}
\end{subfigure}
\hspace{0.05cm}
\begin{subfigure}{0.32\textwidth}
    \centering
    \captionsetup{font=footnotesize}
    \includegraphics[width=0.95\linewidth]{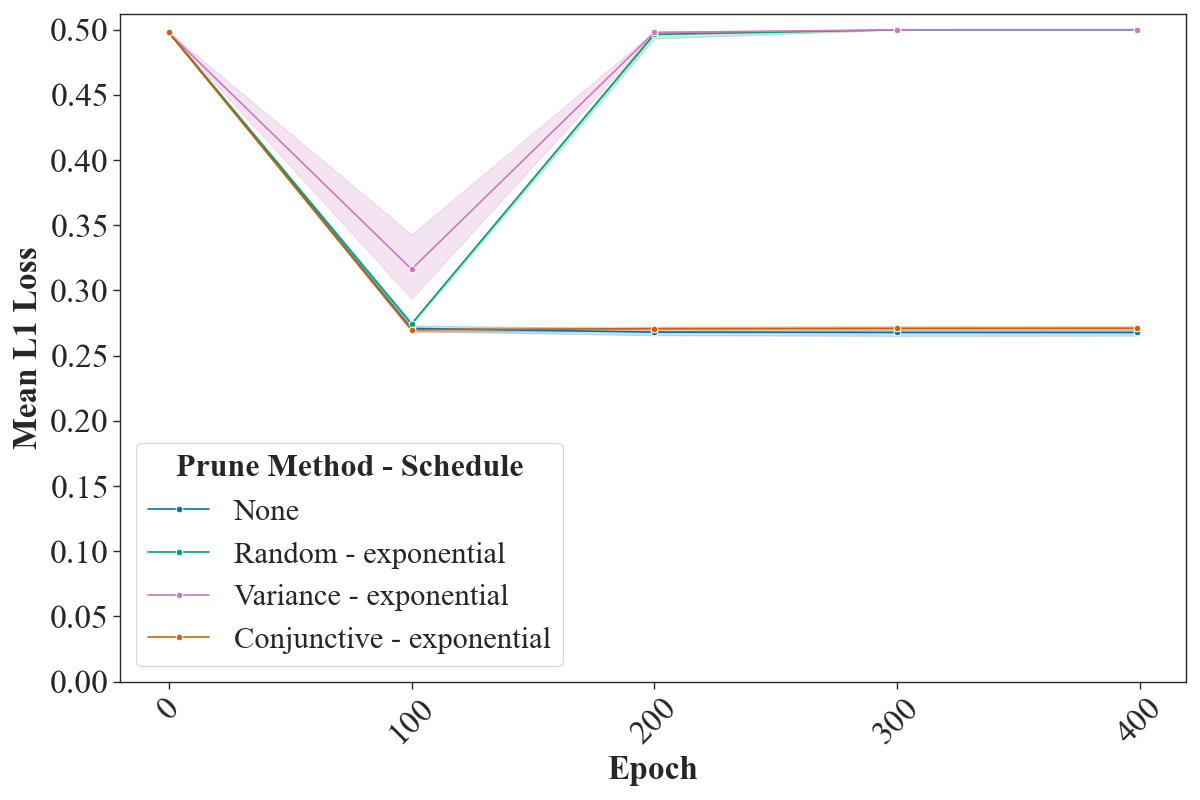}
    \caption{Canonical: Average L1 loss achieved by top-performing pruning schedules per method}
    \label{fig:best-canonical}
\end{subfigure}
\hspace{0.05cm}
\begin{subfigure}{0.32\textwidth}
    \centering
    \captionsetup{font=footnotesize}
    \includegraphics[width=0.95\linewidth]{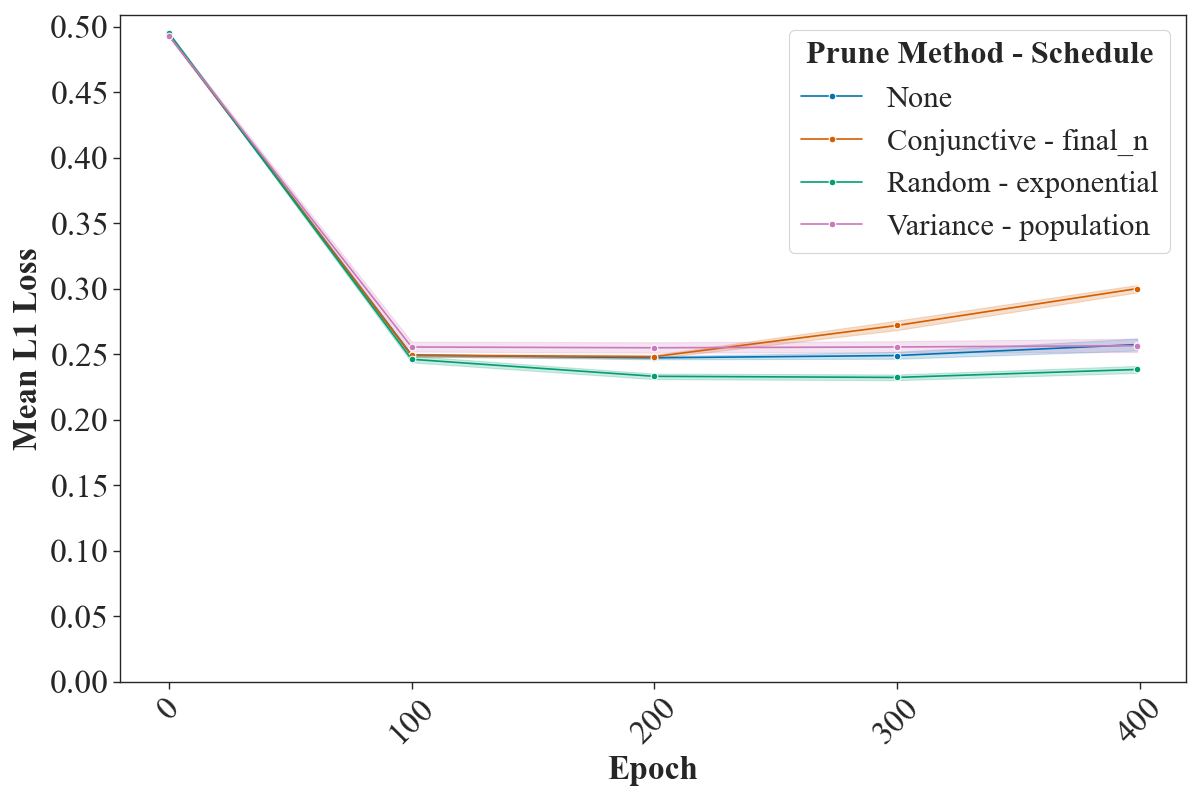}
    \caption{\lipiaesimple-Small: Average L1 loss achieved by top-performing pruning schedules per method}
    \label{fig:best-lipi-small}
\end{subfigure}
  \caption{Comparative performance of best pruning configs across training paradigms: \lipiaesimple (standard and reduced networks) versus canonical AE training}
  \label{fig:best_sweep}
\end{figure*}

Evolution-based pruning of autoencoders can indeed be effective, as demonstrated by our experimental results in Figure~\ref{fig:best-all} and~\ref{fig:best_sweep}. We can see that population-based training using Lipi-AE-S with  the RANDOM pruning method and an exponential pruning schedule performed best, achieving lower loss than the unpruned canonical control while reducing network size. Additionally, the smaller network for Lipi-AE-S was also able to outperform the control, though not to the same extent as the full sized model. This indicates that evolution-based pruning can potentially improve AE performance while decreasing model complexity. However, care must be taken to not start the AE with too small of a capacity, or it will hinder performance.

Figures~\ref{fig:best-lipi} and~\ref{fig:best-canonical} highlight that \lipiaesimple consistently outperforms canonical AE training across all tested pruning methods. This supports the hypothesis that population-based coevolution enhances the autoencoder’s performance and ability to adapt and remain resilient to changes in its architecture. In Figure~\ref{fig:best-canonical}, VARIANCE and RANDOM pruning methods show an initial improvement in performance, followed by a noticeable decline, suggesting that some pruning initially improves performance, but further pruning by these methods leads to over-pruning of the network.

Interestingly, the \MuLexicase pruning method performs comparably to the unpruned variant of canonical AE training, despite having fewer parameters due to pruning. This indicates that \MuLexicase is effective at maintaining performance while reducing network complexity, making canonical AE training more efficient without sacrificing performance. In contrast, the other pruning methods struggled to achieve this balance, underscoring the unique advantages of \MuLexicase in preserving critical connections during training.

Figure~\ref{fig:best-lipi-small} shows that smaller networks experience a decline in performance when pruned, likely due to their reduced latent space capacity compared to the default AE architecture. This observation highlights the importance of starting with an adequate network capacity to maintain performance during pruning.

\subsubsection{What is the impact of pruning mutation operators and pruning schedules on population-based training?}

\begin{figure*}[]
    \captionsetup{font=footnotesize}
\begin{subfigure}{0.32\textwidth}
    \centering
    \captionsetup{font=footnotesize}
    \includegraphics[width=0.95\linewidth]{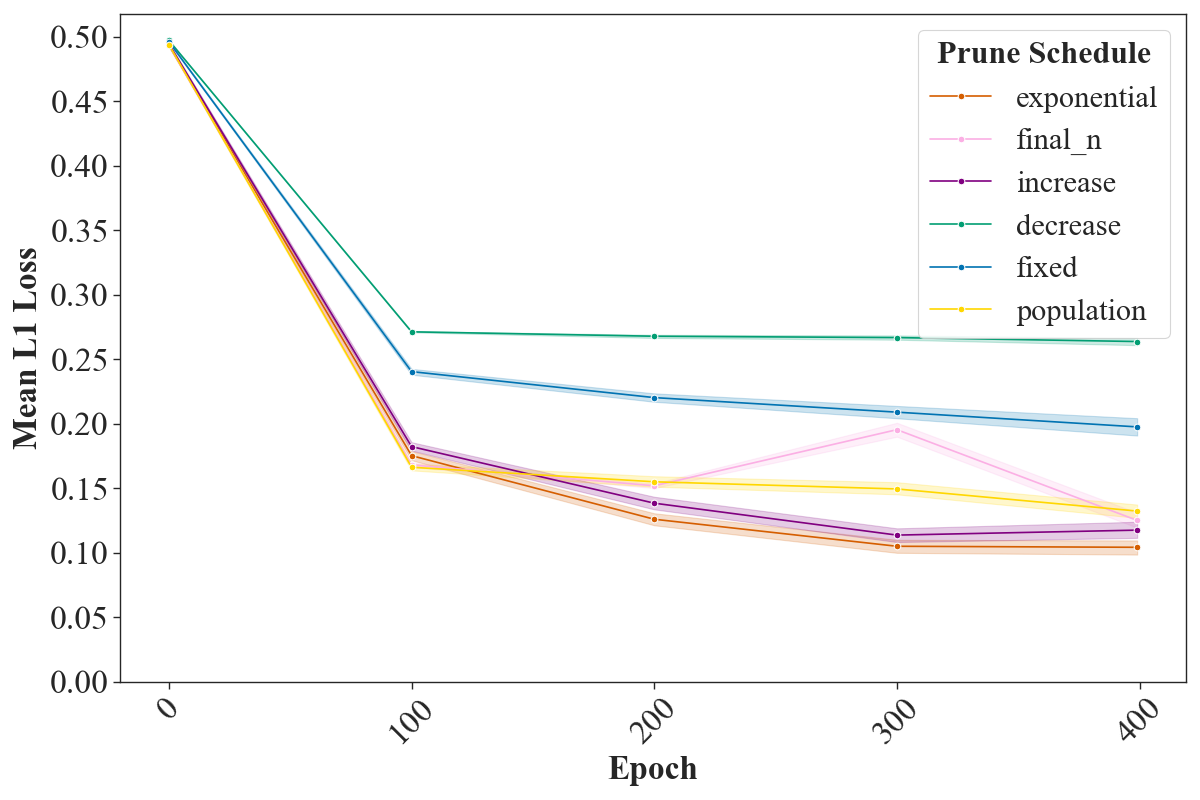}
    \caption{Average L1 loss for RANDOM pruning on all schedules for \lipiaesimple}
    \label{fig:schedule-random}
\end{subfigure}
\hspace{0.05cm}
\begin{subfigure}{0.32\textwidth}
    \centering
    \captionsetup{font=footnotesize}
    \includegraphics[width=0.95\linewidth]{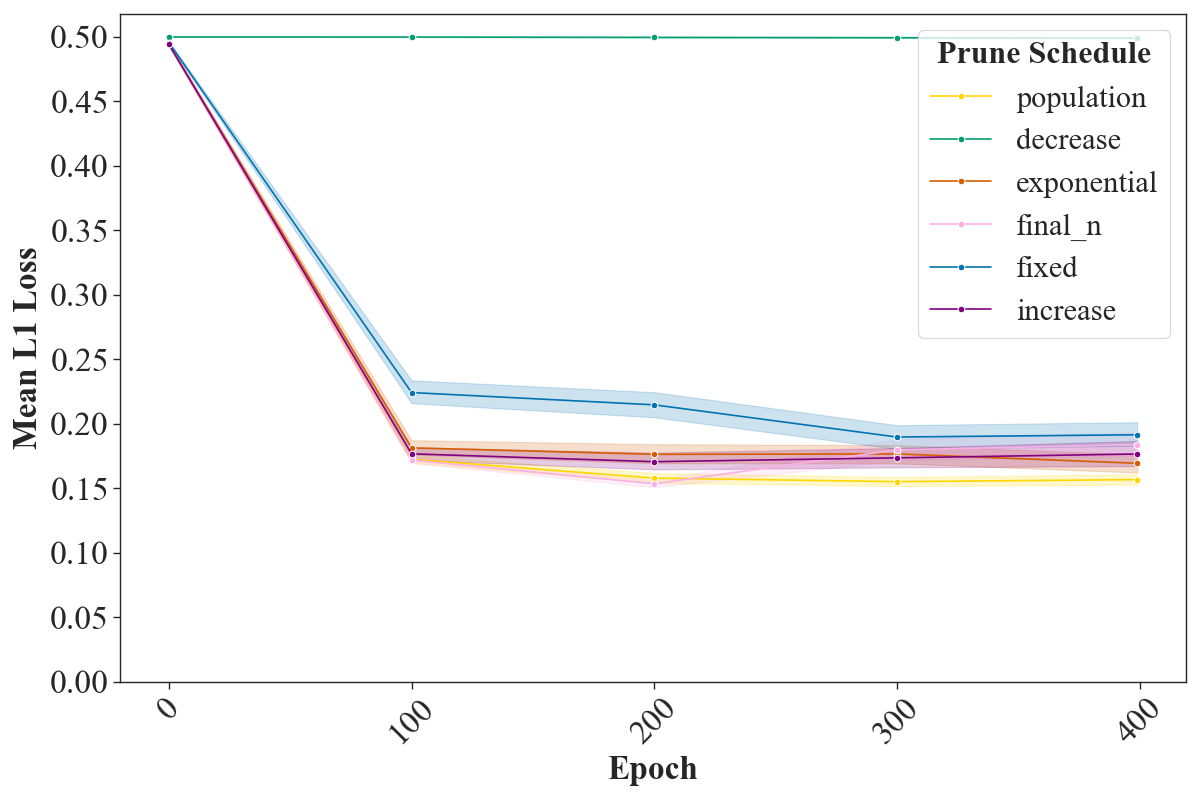}
    \caption{Average L1 loss for VARIANCE pruning on all schedules for \lipiaesimple}
    \label{fig:schedule-activation}
\end{subfigure}
\hspace{0.05cm}
\begin{subfigure}{0.32\textwidth}
    \centering
    \captionsetup{font=footnotesize}
    \includegraphics[width=0.95\linewidth]{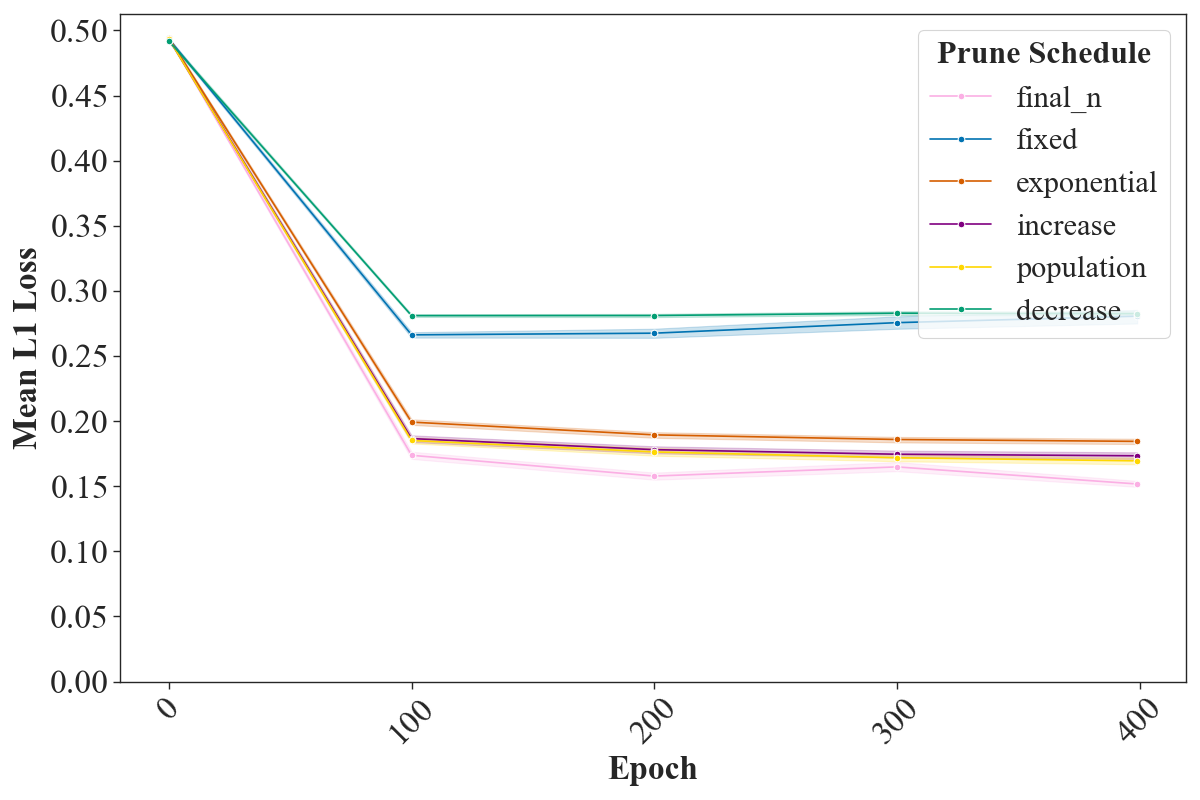}
    \caption{Average L1 loss for CONJUNCTIVE pruning on all schedules for \lipiaesimple}
    \label{fig:schedule-lexi}
\end{subfigure}
  \caption{Pruning Schedule performance for each of the pruning methods when using \lipiaesimple Training}
  \label{fig:pruning_schedule_sweep}
\end{figure*}

Our analysis highlights the complex interplay between pruning methods and scheduling strategies in population-based AE training. 
Figure~\ref{fig:pruning_schedule_sweep} demonstrates that the VARIANCE pruning method performs best when paired with the population-based schedule (Figure~\ref{fig:schedule-activation}). This combination appears to mitigate over-pruning by lowering pruning probabilities at the individual level, which helps preserve network capacity during training. In contrast, VARIANCE pruning under other schedules tends to result in higher mean L1 loss, suggesting that its effectiveness is highly dependent on the schedule used.

The \MuLexicase pruning method exhibits comparable mean L1 loss to VARIANCE across most schedules, with the final-$n$ schedule achieving the lowest loss for \MuLexicase (Figure~\ref{fig:schedule-lexi}). This result indicates that \MuLexicase benefits from schedules that concentrate pruning efforts during the later stages of training, allowing it to maintain efficiency while preserving important parameters. Interestingly, this performance is similar to the best results achieved by VARIANCE with its optimal schedule, suggesting that both methods can be effective under carefully tuned conditions.

The RANDOM pruning method generally outperformed VARIANCE and \MuLexicase across most schedules, with the exponential schedule yielding the lowest overall mean L1 loss. This result highlights the advantages of stochastic exploration in population-based training, as RANDOM pruning introduces greater architectural variability that seems to benefit coevolutionary systems. However, decreasing schedules consistently resulted in higher mean L1 loss across all pruning methods, reinforcing the importance of selecting appropriate pruning schedules to avoid over-pruning.

These findings show that both the pruning method and schedule are important in determining AE performance. Population-based training seems particularly well-suited to stochastic approaches like RANDOM pruning, which uses architectural diversity to enhance resilience and adaptability. In contrast, canonical AE training favors more stable methods like \MuLexicase, which provide consistent performance improvements without relying on population-driven dynamics.

\begin{figure}
    \centering
    \captionsetup{font=footnotesize}
    \includegraphics[width=0.94\linewidth]{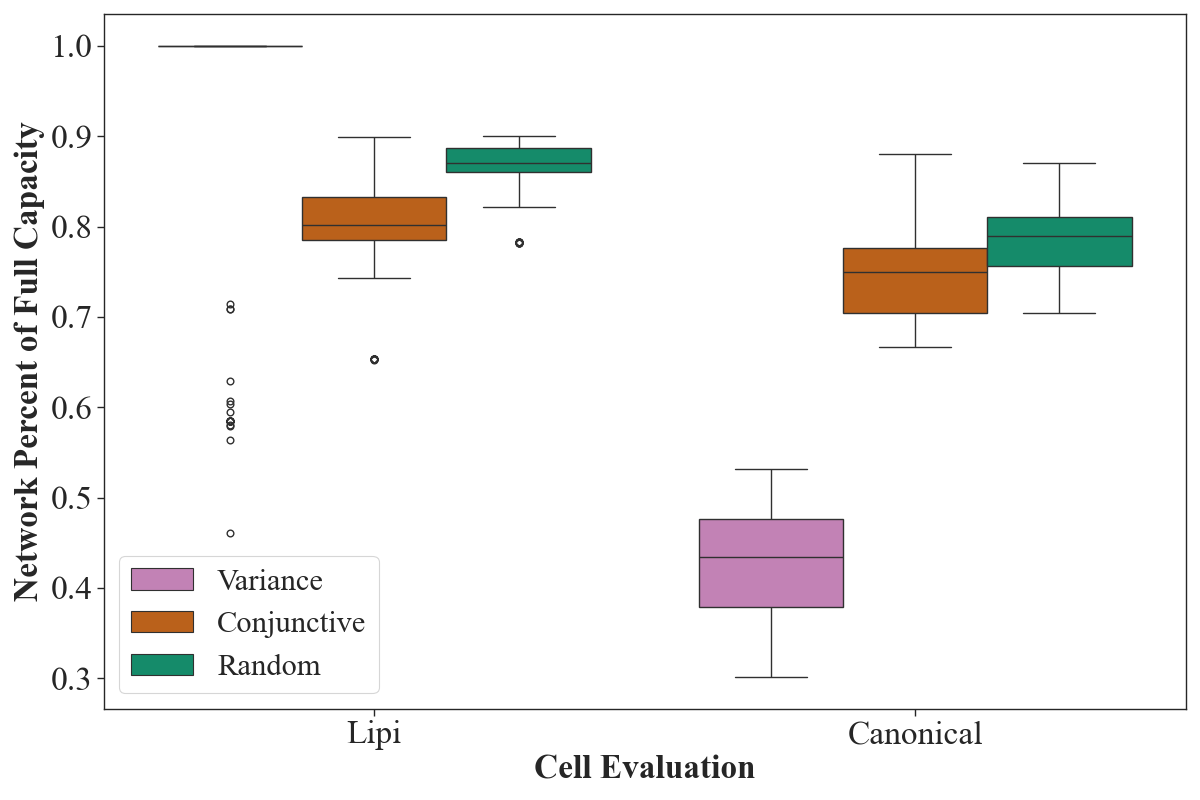}
    \caption{Preserved Percent of Full Network Size post-Training}
    \label{fig:box-capacity}
\end{figure}

\begin{figure}
    \centering
    \captionsetup{font=footnotesize}
    \includegraphics[width=0.94\linewidth]{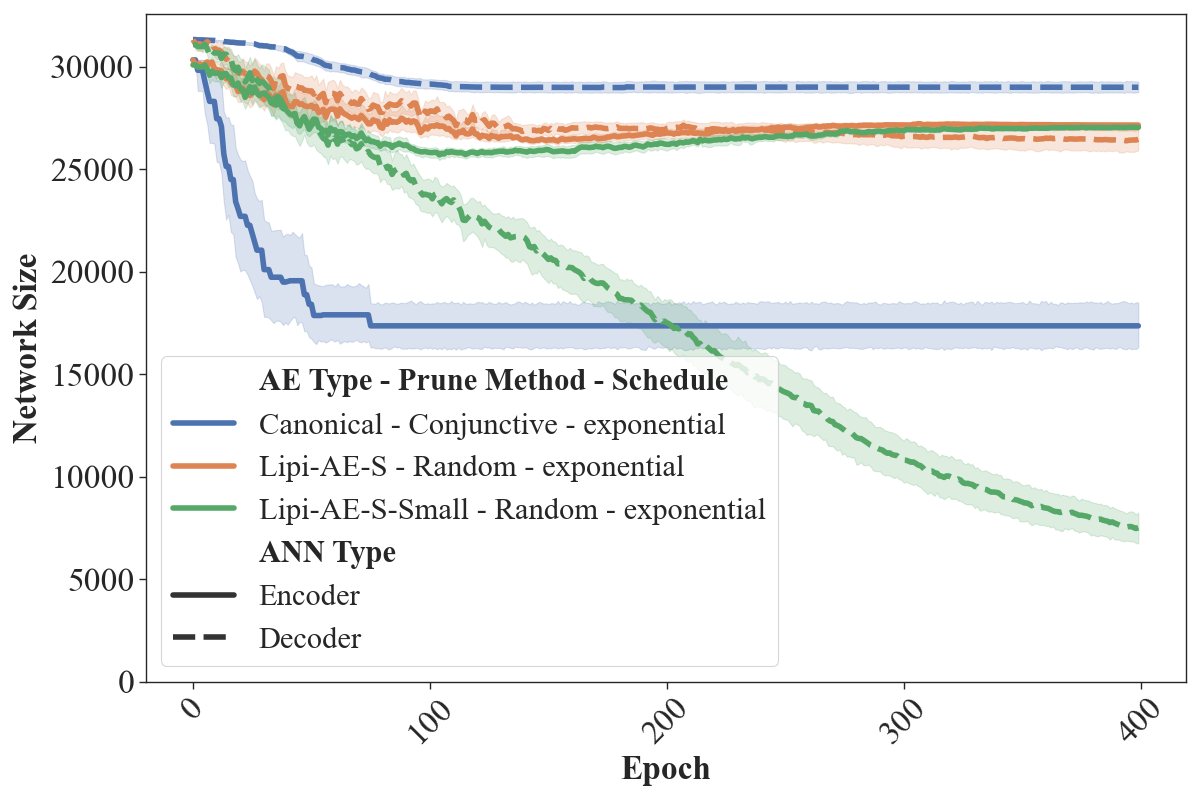}
    \caption{Average Network Size of best pruned individual from \lipiaesimple, \lipiaesimple-Small, and Canonical}
    \label{fig:best-individual}
\end{figure}

\subsubsection{How do the methods compare in terms of preserved-percentage of the original network?}
The preserved percentage of the original network varied across pruning methods and training approaches. 
Figure~\ref{fig:box-capacity} provides a comprehensive view of the final preserved percentages across different pruning methods and training approaches.
For canonical AE training, CONJUNCTIVE pruning achieved the best balance between performance and network size reduction. This method preserved approximately 70-80\% of the original network while maintaining comparable performance to the unpruned baseline. In contrast, RANDOM pruning preserved a slightly larger percentage of the network (around 80\%) but resulted in significantly higher loss, highlighting its ineffectiveness for canonical training. VARIANCE pruning, while more aggressive in reducing network size (preserving only 40-50\% of the original network), also led to substantially higher loss, suggesting that over-pruning compromises performance in canonical training.

For population-based training (\lipiaesimple), RANDOM pruning emerged as the most effective method, delivering strong performance while preserving approximately 88\% of the original network. CONJUNCTIVE pruning was more aggressive, reducing the preserved percentage to around 80\%, but achieved a lower overall performance compared to the RANDOM method. Interestingly, VARIANCE pruning, despite expectations from the canonical training, preserved the largest percentage of the original network, pruning the fewest nodes (about 99\% remaining), but only achieved similar performance to the \MuLexicase method. This result suggests that the population-based approach mitigated over-pruning by favoring models with better performance during selection. However, this also tempered pruning overall, limiting its effectiveness for VARIANCE while benefiting methods like \MuLexicase and RANDOM by removing more unnecessary nodes without compromising accuracy.

Figure~\ref{fig:best-individual} reveals architectural evolution patterns between encoder-decoder pairs across training paradigms. In \lipiaesimple with RANDOM-exponential pruning, we observe symmetrical parameter retention in both encoder and decoder, suggesting balanced architectural optimization through population-driven exploration. The \lipiaesimple-Small variant maintains encoder dimensions comparable to the base model but achieves higher decoder compression, likely due to reduced latent space dimensionality creating a bottleneck that constrains decoder complexity requirements.
Canonical AE training with \MuLexicase exhibits stark asymmetry, demonstrating decoder resilience to encoder over-pruning. Unlike population-based approaches, canonical training lacks selection pressure to temper aggressive encoder reduction, forcing the decoder to preserve critical reconstruction pathways.

Overall, the study establishes that evolutionary pruning is not only a viable strategy for optimizing autoencoders but also one that requires careful consideration of initial network capacity, pruning schedules, and training paradigms. Population-based training seems effective in leveraging architectural diversity to enhance adaptability and resilience, and we see that pruning too much too early can be detrimental to model performance.

\subsection{Limitations}
\label{sec:discussion}

While our study provides valuable insights into the application of spatial cooperative coevolution and pruning techniques for autoencoder training, several important limitations constrain the generalizability of our findings. Our experiments primarily investigated a
few problem variations in a single domain, potentially overlooking more complex or nuanced behaviors of our proposed method. In addition, we examined
only a limited number of AE architectures and sizes, as well as hyperparameters. Our approach also examined only a subset of potential pruning methods, which may not fully capture the diverse strategies available for neural network compression. These constraints underscore the need for more multi-domain studies that can validate and extend our findings across a broader range of scenarios and computational contexts.

\section{Conclusions \& Future Work}
\label{sec:conclusions}

In this study, we introduce novel mutation operators for pruning autoencoders (AEs) based on their activation behavior. Our comparative analysis encompasses both population-based and canonical AE training methods. The results demonstrate that canonical AE training exhibits greater susceptibility to pruning effects compared to population-based training using Lipi-AE-S.
Additionally, we demonstrate that the effectiveness of pruning methods varies across training approaches.
For \lipiaesimple pruning, random weights perform best. For canonical AE pruning \lexicase performs best. We postulate that the population of AEs can select the best AEs based on pruning random weights. For canonical AE training, pruning weights have a high variance and the use of \lexicase has a more stable performance. Moreover, we find that pruning too early does not improve performance. Pruning schedules that prune with higher probability later in the training, such as the exponential or final-$n$ schedule explored in this paper, have the highest performance.

Future work will investigate more problems in different problem
domains, such as images and text. We will also investigate the dynamics
of the encoder-decoder pairs similarity and convergence to improve the
understanding of AE convergence during training. Furthermore, we will
investigate the impact of the AE architecture. There are additional pruning operators such as Low-Rank Matrix Approximation to explore, as well as heterogeneous pruning operators on the different spatial nodes and fully removing nodes. Moreover, more fine-grained methods for measuring computational effort, e.g. FLOPS can be used to simply counting parameters. Finally, we will expand with theoretical analysis.

\begin{acks}
This research was partially funded by the Universidad de Málaga, Spain, grant B1-2022\_18, and by QUAL21-010 UMA. The authors thank the Supercomputing and Bioinformatics center at the UMA for their computer resources and assistance. 

DISTRIBUTION STATEMENT A.
Approved for public release.
Distribution is unlimited.
This material is based upon work supported by the Department of the Air Force under Air Force Contract No.~FA8702-15-D-0001.
Any opinions, findings, conclusions or recommendations expressed in this material are those of the author(s) and do not necessarily reflect the views of the Department of the Air Force.
\end{acks}



\balance
\bibliographystyle{ACM-Reference-Format}
\bibliography{references,rw}


\appendix

\section{Additional Figures and Results}
\label{sec:results_appendix}

\begin{figure}
    \centering
    \captionsetup{font=footnotesize}
    \includegraphics[width=0.88\linewidth]{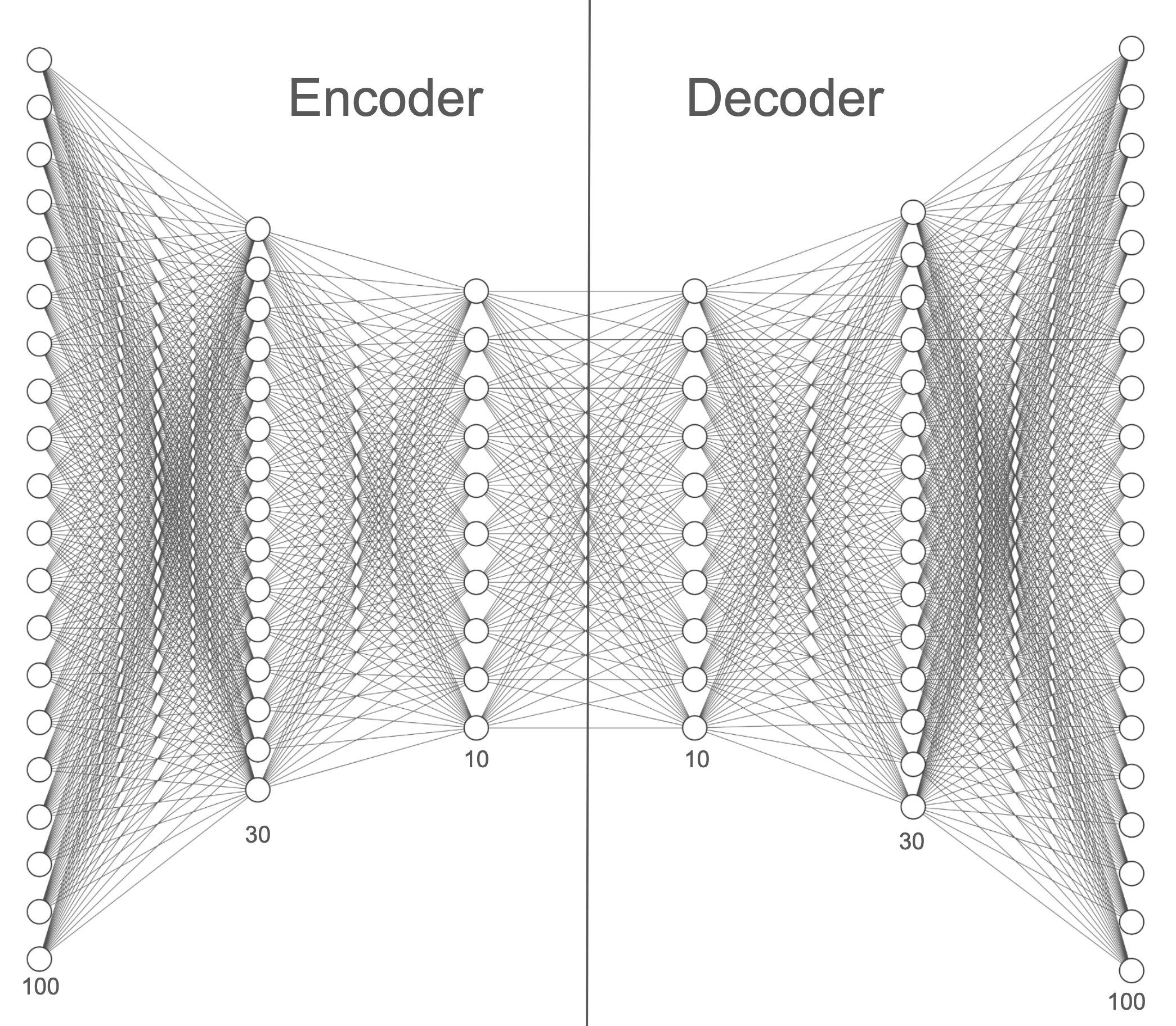}
    \caption{Architecture diagram of our default autoencoders}
    \label{fig:base_ann_arch}
\end{figure}

The ANNs in this study were implemented using PyTorch, a flexible and efficient deep learning framework. Figure \ref{fig:base_ann_arch} illustrates the default architecture of the autoencoders used in our experiments. More details can be found in our github repository, linked above.

\begin{figure}
    \centering
    \captionsetup{font=footnotesize}
    \includegraphics[width=0.88\linewidth]{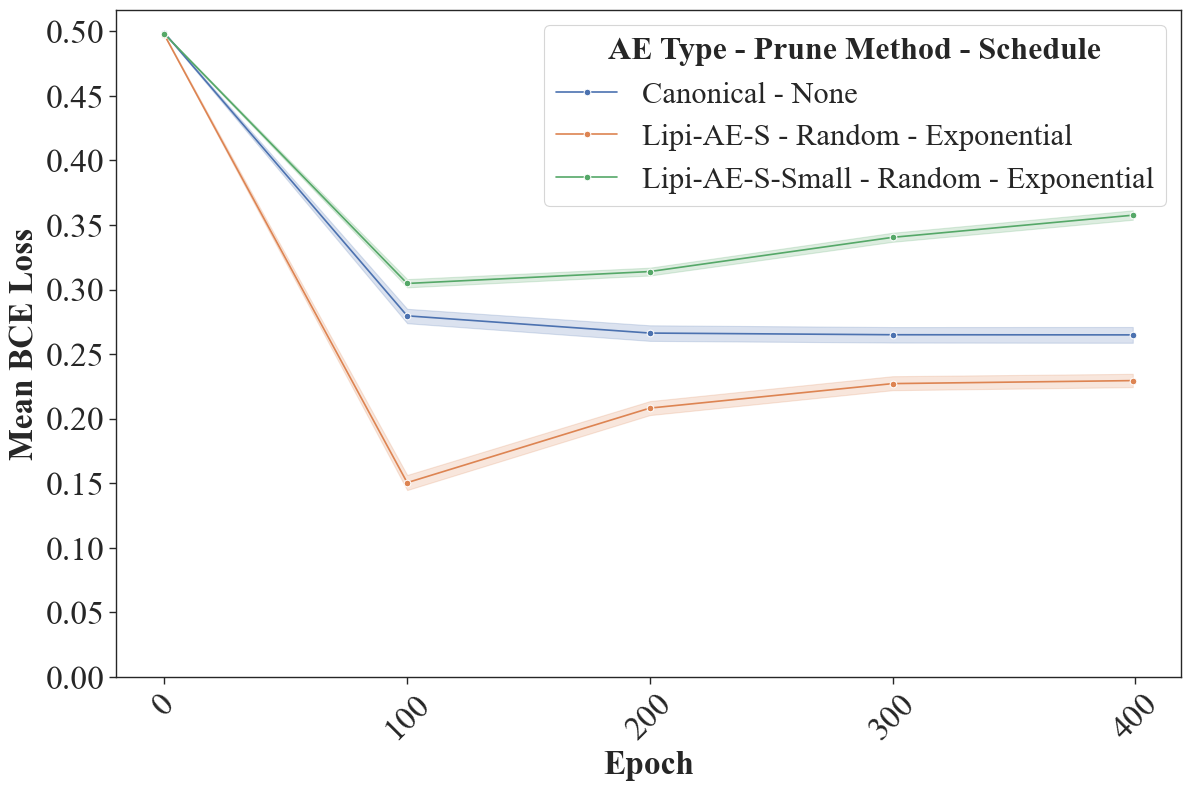}
    \caption{Best Pruning Method/Schedule combos when trained with BCE loss}
    \label{fig:bce_best_of_best}
\end{figure}

\begin{figure}
    \centering
    \captionsetup{font=footnotesize}
    \includegraphics[width=0.88\linewidth]{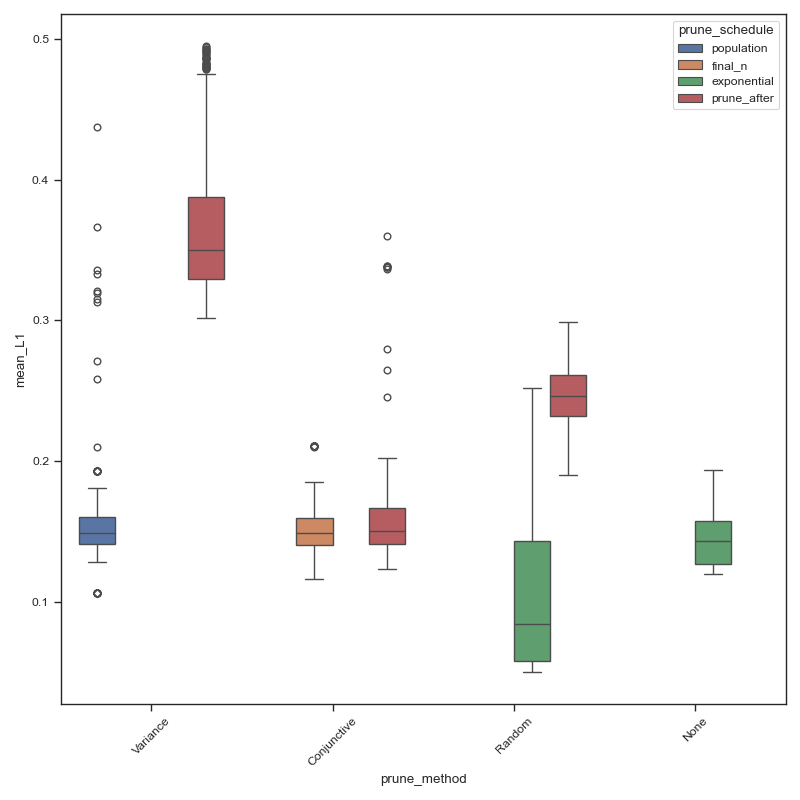}
    \caption{Best performing pruning method/schedules for Lipi-AE-S, compared to pruning after training}
    \label{fig:best-lipi-box}
\end{figure}

Figure \ref{fig:bce_best_of_best} shows the loss progression across epochs for the optimal pruning method and schedule combinations applied to \lipiaesimple, \lipiaesimple-Small, and Canonical training setups, using Binary Cross Entropy (BCE) loss instead of the L1 loss employed in the main study. Overall, these combinations demonstrate inferior performance compared to those using L1 loss.
Figure ~\ref{fig:best-lipi-box} provides a comparative analysis of the top-performing methods and schedules for \lipiaesimple, benchmarked against the typical post-training pruning. 
\section{Ablation Study}
\label{sec:ablation-study}

In this section, we present an ablation study to analyze the impact of key design choices in our method.  
Specifically, we investigate the effects of different population sizes (1, 2, 5, 10, 20), network architectures (Small Autoencoder, Standard Autoencoder, Large Autoencoder, and Denoising Autoencoder), and cell evaluation methods (epoch-node, canonical, all vs all, and simple). Each configuration has been independently tested across 30 experimental runs and mean L1 and minimum selection losses are evaluated.

\subsection{Population Size}

\begin{figure}
    \centering
    \captionsetup{font=footnotesize}
    \includegraphics[width=0.88\linewidth]{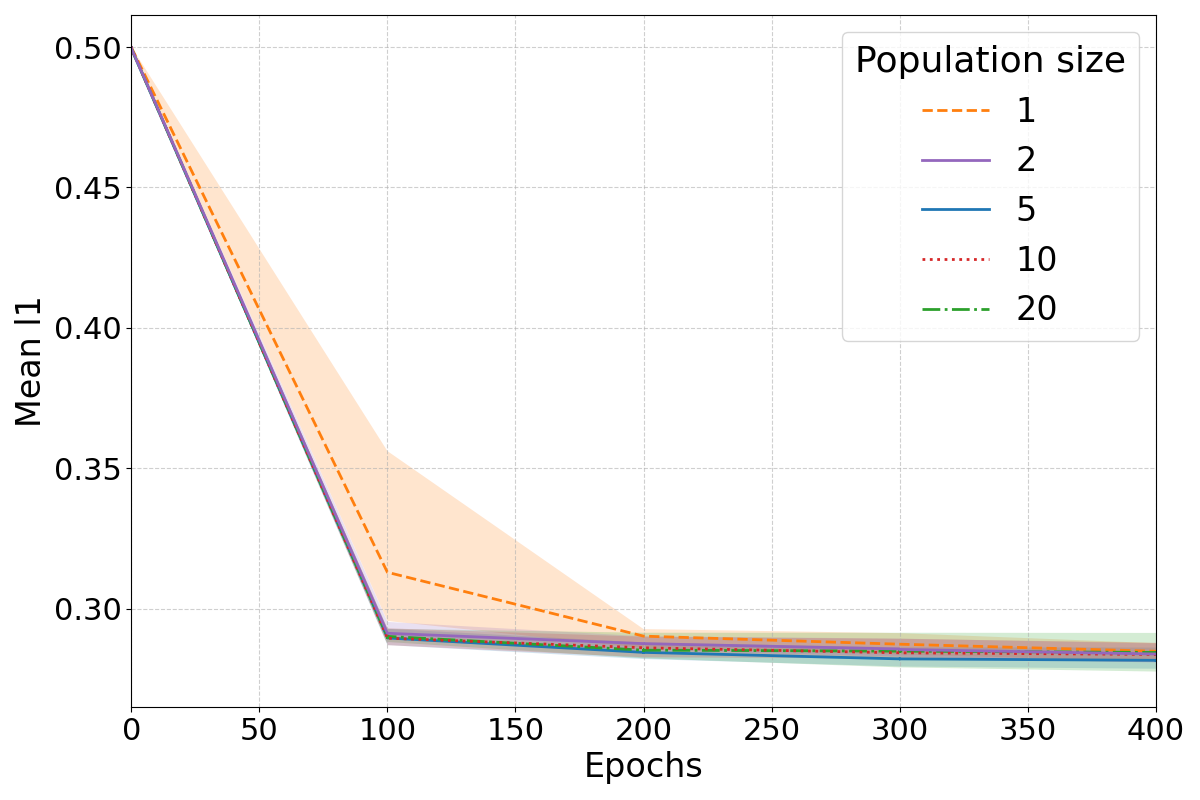}
    \caption{Mean L1 loss across epochs for different population sizes. }
    \label{fig:mean_l1_loss_population_size}
\end{figure}

\begin{figure}
    \centering
    \captionsetup{font=footnotesize}
    \includegraphics[width=0.88\linewidth]{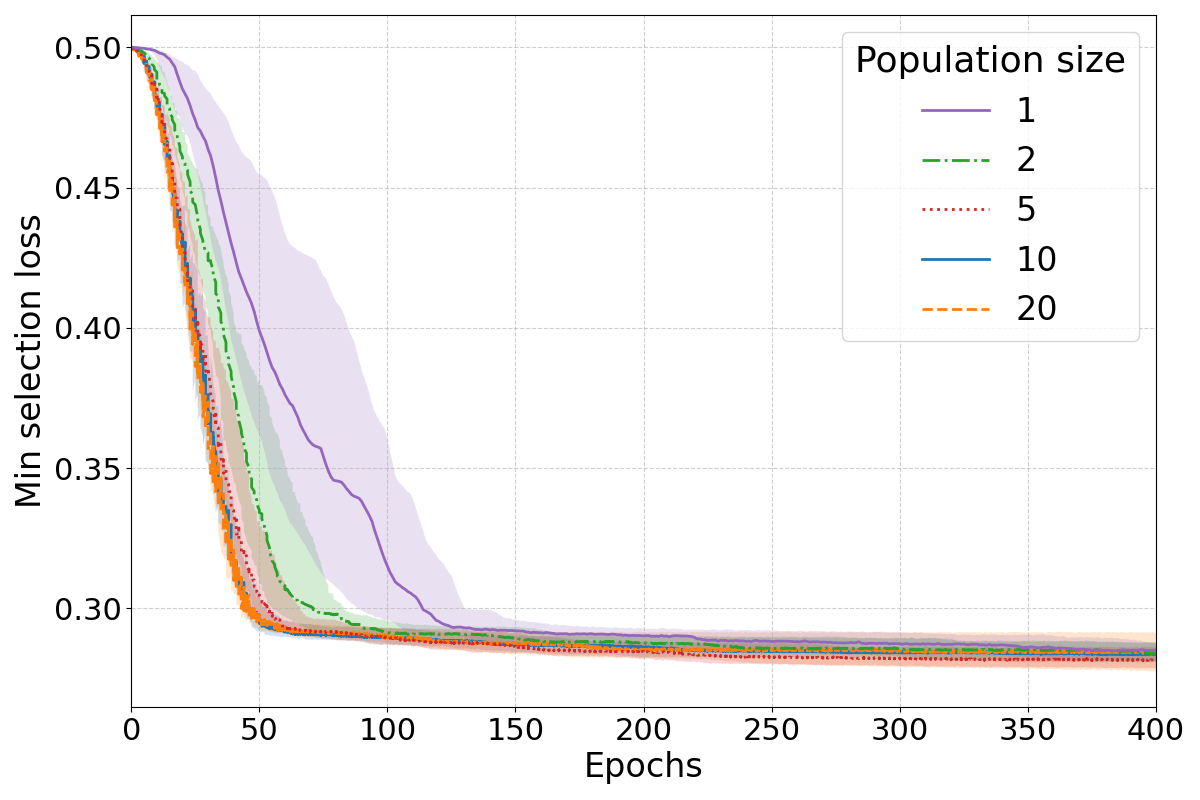}
    \caption{Minimum selection loss across epochs for different population sizes. }
    \label{fig:min_selection_loss_population_size}
\end{figure}

This subsection analyzes the impact of population size on the performance of the proposed method. Figures~\ref{fig:mean_l1_loss_population_size} and~\ref{fig:min_selection_loss_population_size} show the evolution of mean L1 loss and minimum selection loss results, respectively.

The results in Figure~\ref{fig:mean_l1_loss_population_size} highlight that larger populations, such as 10 and 20, achieve lower mean L1 loss and reduced variance compared to smaller sizes. This suggests that larger populations benefit from greater diversity, enabling more effective exploration and optimization of the solution space. Although, the marginal gains diminish for population sizes beyond 10, suggesting a trade-off between performance and computational cost.

The results in Figure~\ref{fig:min_selection_loss_population_size} show that larger populations achieve a steeper decline in minimum selection loss during the initial epochs, indicating faster identification of high-quality solutions. Increased diversity in larger populations improves exploration and contributes to more robust performance, with smaller populations showing higher variance and less stability. As it has been shown when evaluating mean L1 loss, the benefits of using populations with larger sizes diminish beyond a certain size.

\subsection{ANN Architectures}

\begin{figure}
    \centering
    \captionsetup{font=footnotesize}
    \includegraphics[width=0.88\linewidth]{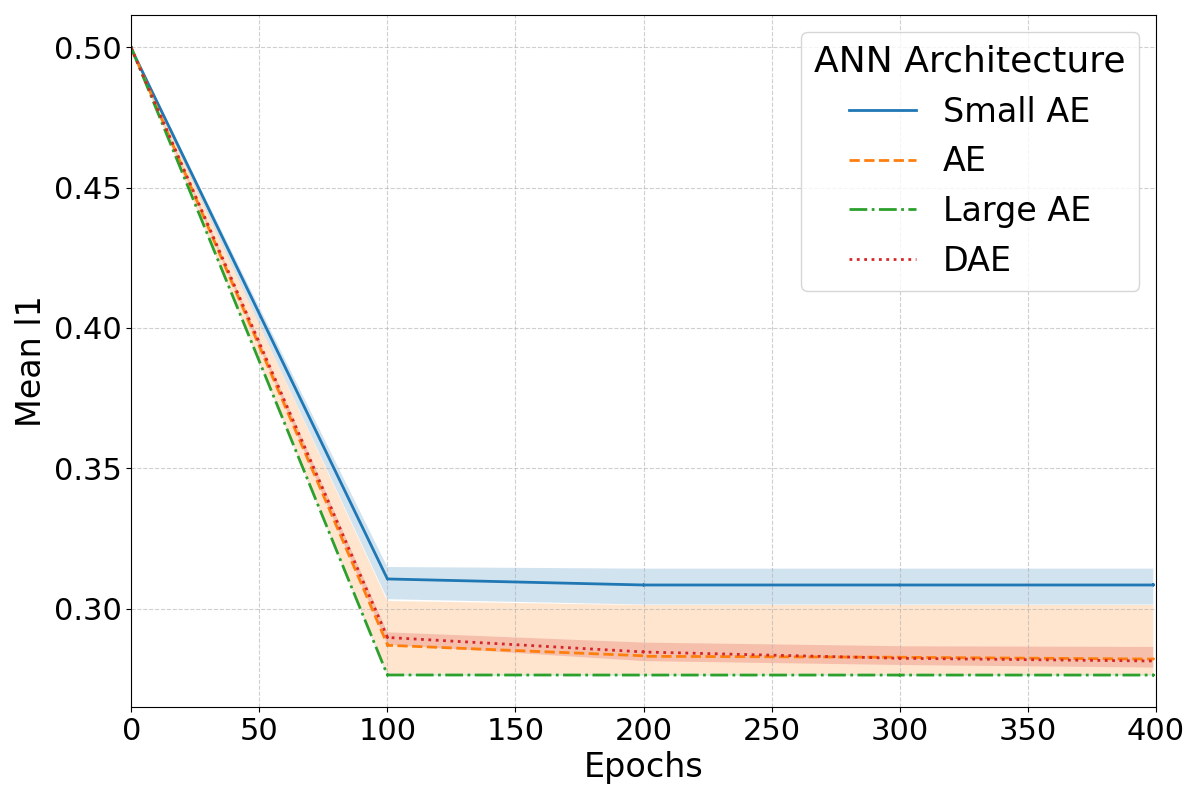}
    \caption{Mean L1 loss across epochs for different ANN architectures.}
    \label{fig:mean_l1_loss_ann_architecture}
\end{figure}

\begin{figure}[]
    \centering
    \captionsetup{font=footnotesize}
    \includegraphics[width=0.88\linewidth]{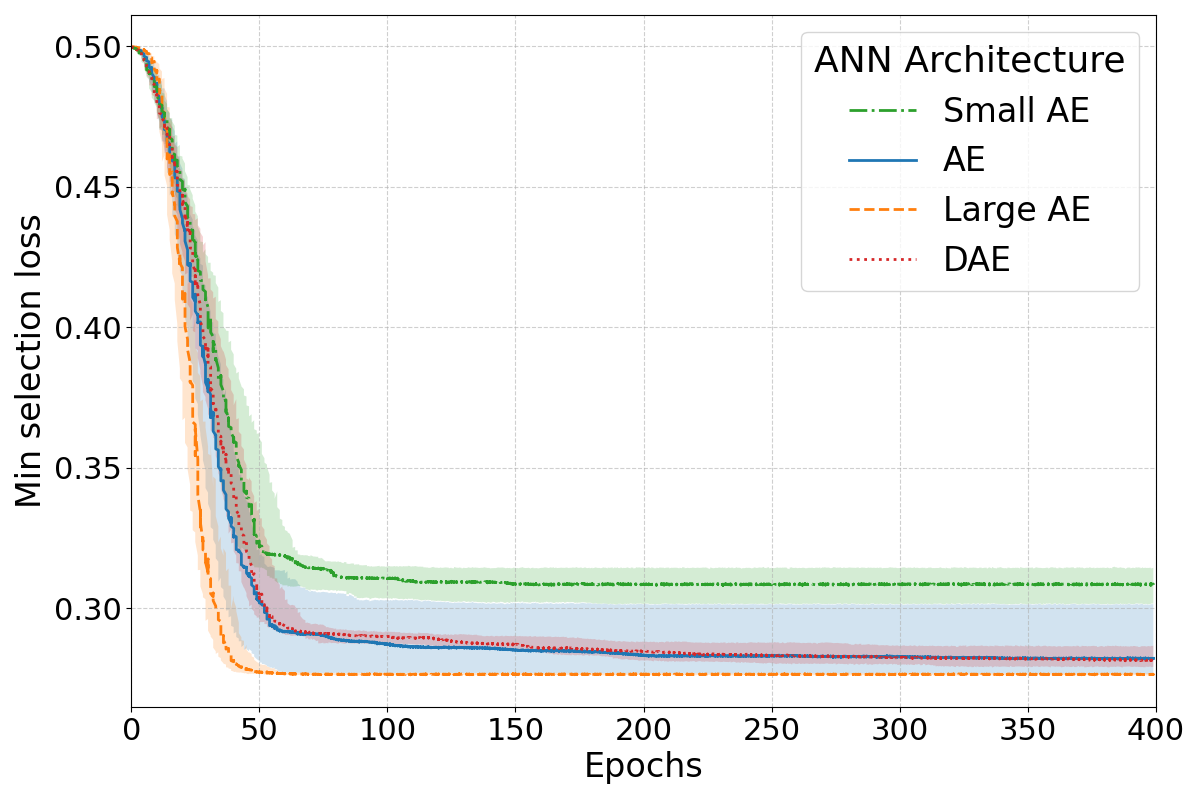}
    \caption{Minimum selection loss across epochs for different ANN architectures.}
    \label{fig:min_selection_loss_ann_architecture}
\end{figure}

This subsection analyzes the impact of different ANN architectures on the performance of the proposed method. The results, presented in Figures~\ref{fig:mean_l1_loss_ann_architecture} and~\ref{fig:min_selection_loss_ann_architecture}, reveal insights into how architectural choices and sizes affect performance.

Figure~\ref{fig:mean_l1_loss_ann_architecture} shows that the Large AE consistently achieves the lowest mean L1 loss compared to the other architectures. In contrast, the Small AE exhibits higher mean L1 loss and variance, reflecting its limited capacity to address this problem effectively. AE and DAE provide similar competitive results. Similar trend is shown in Figure~\ref{fig:min_selection_loss_ann_architecture}, i.e., Large AE achieves the steepest decline in minimum selection loss during the initial epochs. Besides, the variance in minimum selection loss is more pronounced in smaller architectures, emphasizing their instability during training.

\subsection{Cell Evaluation Methods}

\begin{figure}[]
    \centering
    \captionsetup{font=footnotesize}
    \includegraphics[width=0.88\linewidth]{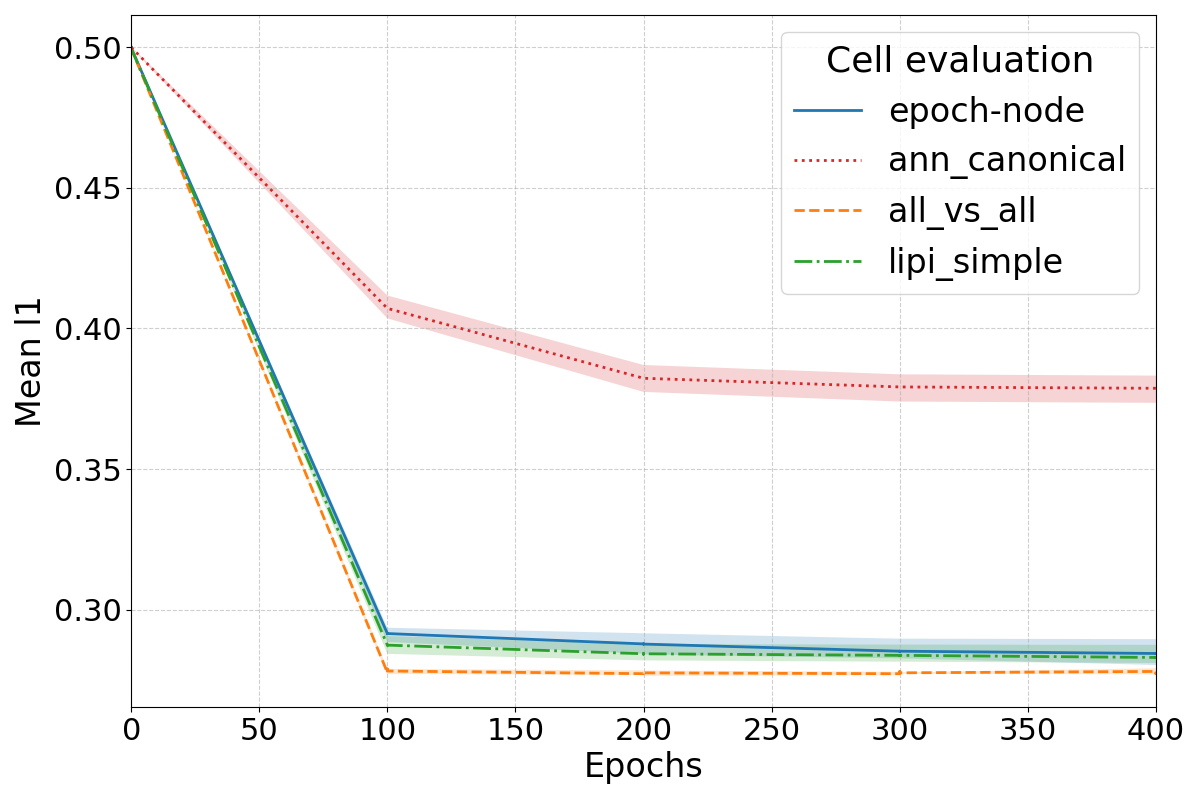}
    \caption{Mean L1 loss across epochs for different cell evaluation methods. }
    \label{fig:mean_l1_loss_cell_evaluation}
\end{figure}

\begin{figure}[]
    \centering
    \captionsetup{font=footnotesize}
    \includegraphics[width=0.88\linewidth]{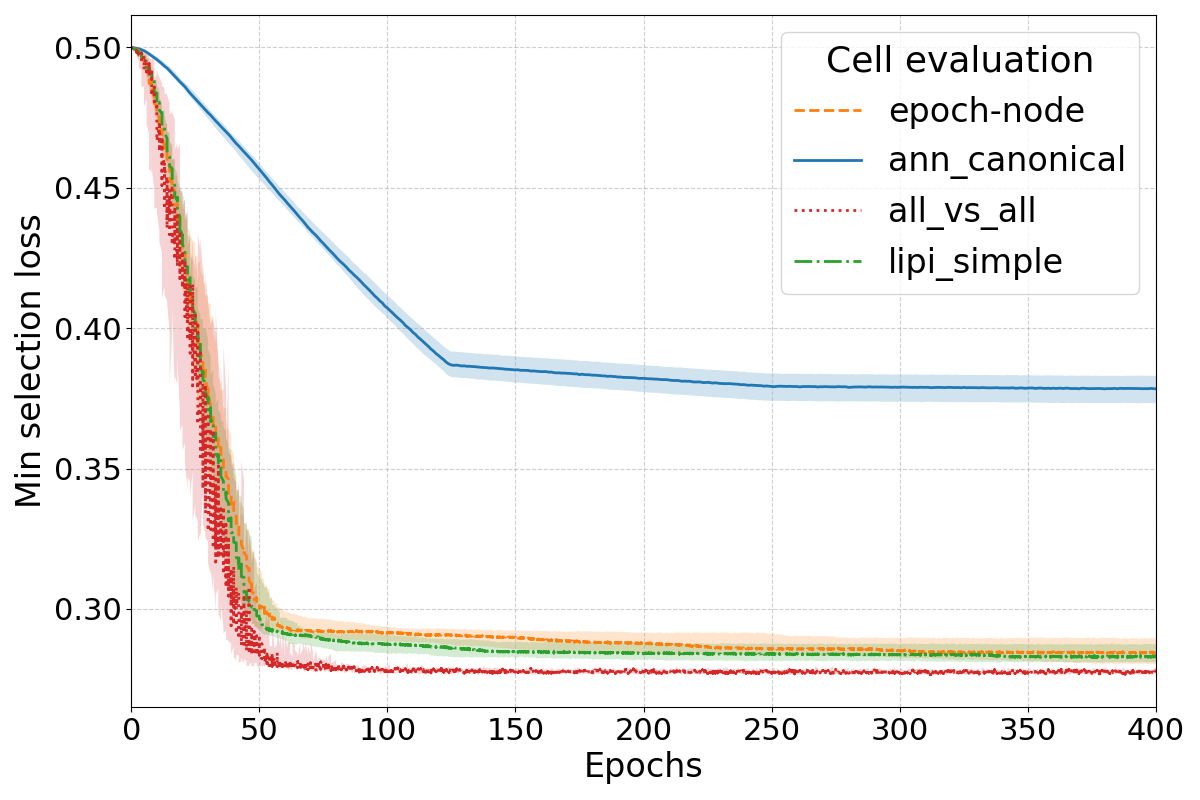}
    \caption{Minimum selection loss across epochs for different cell evaluation methods.}
    \label{fig:min_selection_loss_cell_evaluation}
\end{figure}

This subsection analyzes the impact of different cell evaluation appraches on the performance of the proposed method. The results, presented in Figures~\ref{fig:mean_l1_loss_cell_evaluation} and~\ref{fig:min_selection_loss_cell_evaluation}, reveal insights into how architectural choices affect performance.

Figure~\ref{fig:mean_l1_loss_cell_evaluation} shows that the all vs all cell evaluation method achieve the lowest mean L1 loss, indicating superior performance compared to the other methods. However, it requires higher computational costs than the other methods. 
The ann canonical method exhibits the highest mean L1 loss and variance, reflecting its limited ability to effectively evaluate solutions. 
Lipi simple and epoch-node cell evaluation methods provided competitive mean L1 loss results while requiring lower number of operations.
Figure~\ref{fig:min_selection_loss_cell_evaluation} shows that all cell evaluation methods show a sharp min selection loss reduction in the first 50~training epochs, but the ann canonical method. Among the most competitive methods, the main difference is that all vs all is able to get lower selection losses than epoch-node and lipi simple. Overall, the results suggest that the choice of cell evaluation method has a significant impact on both convergence speed and final performance, with lipi simple and epoch-node providing the reliable outcomes with competitive compuational costs.

\section{Training vs. Test Loss}
\label{ap:training-vs-test-loss}

\begin{figure}[]
    \centering
    \captionsetup{font=footnotesize}
    \includegraphics[width=0.88\linewidth]{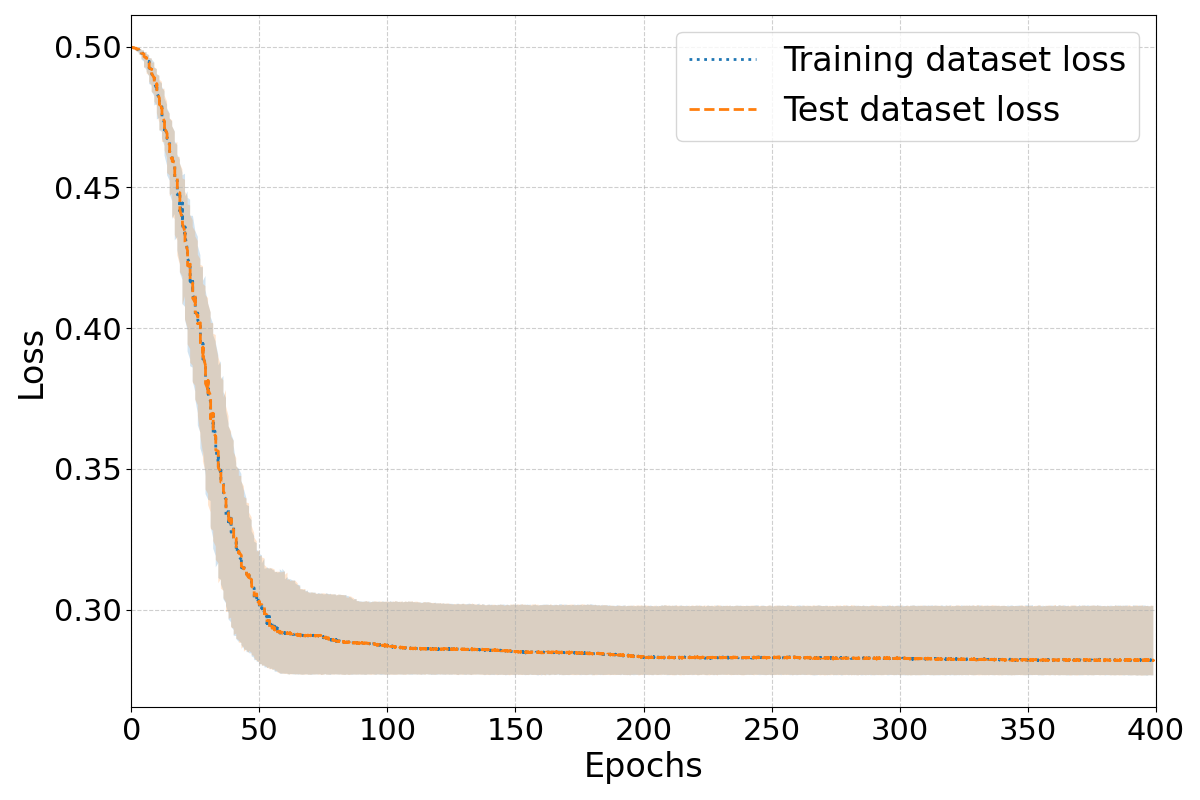}
    \caption{Evolution of test and training dataset losses.}
    \label{fig:test-vs-training}
\end{figure}

In this paper, we report test dataset loss as a representative metric when discussing the results. This choice is due to the losses calculated on the training and test datasets exhibit highly similar behavior. Figure~\ref{fig:test-vs-training} provides an example of the test and training loss curves for the experiment involving the training of an AE using \lipi with a population size of five individuals. The figure demonstrates that the trajectories of the test and training losses are nearly indistinguishable.

\end{document}